\documentclass[runningheads]{llncs}

\usepackage{eccv}

\usepackage{eccvabbrv}

\usepackage{graphicx}
\usepackage{booktabs}

\usepackage{bbding}
\usepackage{adjustbox}
\usepackage{makecell}
\usepackage{multirow} 
\usepackage{wrapfig,blindtext}
\usepackage{amssymb}
\usepackage{color,xcolor,colortbl}
\usepackage[accsupp]{axessibility}  

\usepackage[pagebackref,breaklinks,colorlinks,citecolor=eccvblue]{hyperref}

\usepackage{orcidlink}

\begin{document}

\title{LLMGA: Multimodal Large  Language Model based Generation Assistant} 


\author{	Bin Xia\inst{1}, Shiyin Wang\inst{2}, Yingfan Tao\inst{2}, Yitong Wang\inst{2}, and Jiaya Jia\inst{1} }
\institute{ The Chinese University of Hong Kong \and ByteDance Inc \\\href{https://llmga.github.io/}{https://llmga.github.io/}}




\maketitle

\begin{abstract}
In this paper, we introduce a Multimodal Large Language Model-based Generation Assistant (LLMGA), leveraging the vast reservoir of knowledge and proficiency in reasoning, comprehension, and response inherent in Large Language Models (LLMs) to assist users in image generation and editing. Diverging from existing approaches where Multimodal Large Language Models (MLLMs) generate fixed-size embeddings to control Stable Diffusion (SD), our LLMGA provides a detailed language generation prompt for precise control over SD. This not only augments LLM context understanding but also reduces noise in generation prompts, yields images with more intricate and precise content, and elevates the interpretability of the network. To this end, we curate a comprehensive dataset comprising prompt refinement, similar image generation, inpainting \& outpainting, and instruction-based editing. Moreover, we propose a two-stage training scheme. In the first stage, we train the MLLM to grasp the properties of image generation and editing, enabling it to generate detailed prompts. In the second stage, we optimize SD to align with the MLLM's generation prompts. Additionally, we propose a reference-based restoration network to alleviate texture, brightness, and contrast disparities between generated and preserved regions during inpainting and outpainting. Extensive results show that LLMGA has promising generation and editing capabilities and can enable more flexible and expansive applications in an interactive manner.
\end{abstract}

\vspace{-2mm}
\section{Introduction}
\label{sec:intro}

Artificial Intelligence Generated Content (AIGC) has witnessed remarkable advancements, particularly propelled by the evolution of large language models (LLMs)~\cite{brown2020language,palm,llama} for text generation and diffusion models (DMs)~\cite{DDPM2,LDM} for image generation.
LLMs, in particular, have received considerable acclaim for their exceptional ability to comprehend, reason, make decisions, possess extensive knowledge, and generate text with unparalleled accuracy and fluency.

Recent studies have begun delving deeper into Multimodal Large Language Models (MLLMs)~\cite{flamingo,minigpt4} built upon LLMs, aiming to empower LLMs to comprehend inputs extending beyond text. For example, BLIP-2~\cite{BLIP-2} and LLaVA~\cite{LLAva} employ visual encoders to transform images into input embeddings, enabling them to be used as prompts alongside text input for the LLM, thus achieving  LLMs with the visual modality. Furthermore, recent works focused on extending the capabilities of LLMs to generate multimodal outputs. For example, GILL~\cite{GILL} involves instructing LLMs to predict fixed-size visual embeddings aligned with CLIP~\cite{CLIP} space to control the Stable Diffusion~\cite{LDM} (SD) for image generation.

\begin{figure*}[t]
	\centering
	\includegraphics[height=5.7cm]{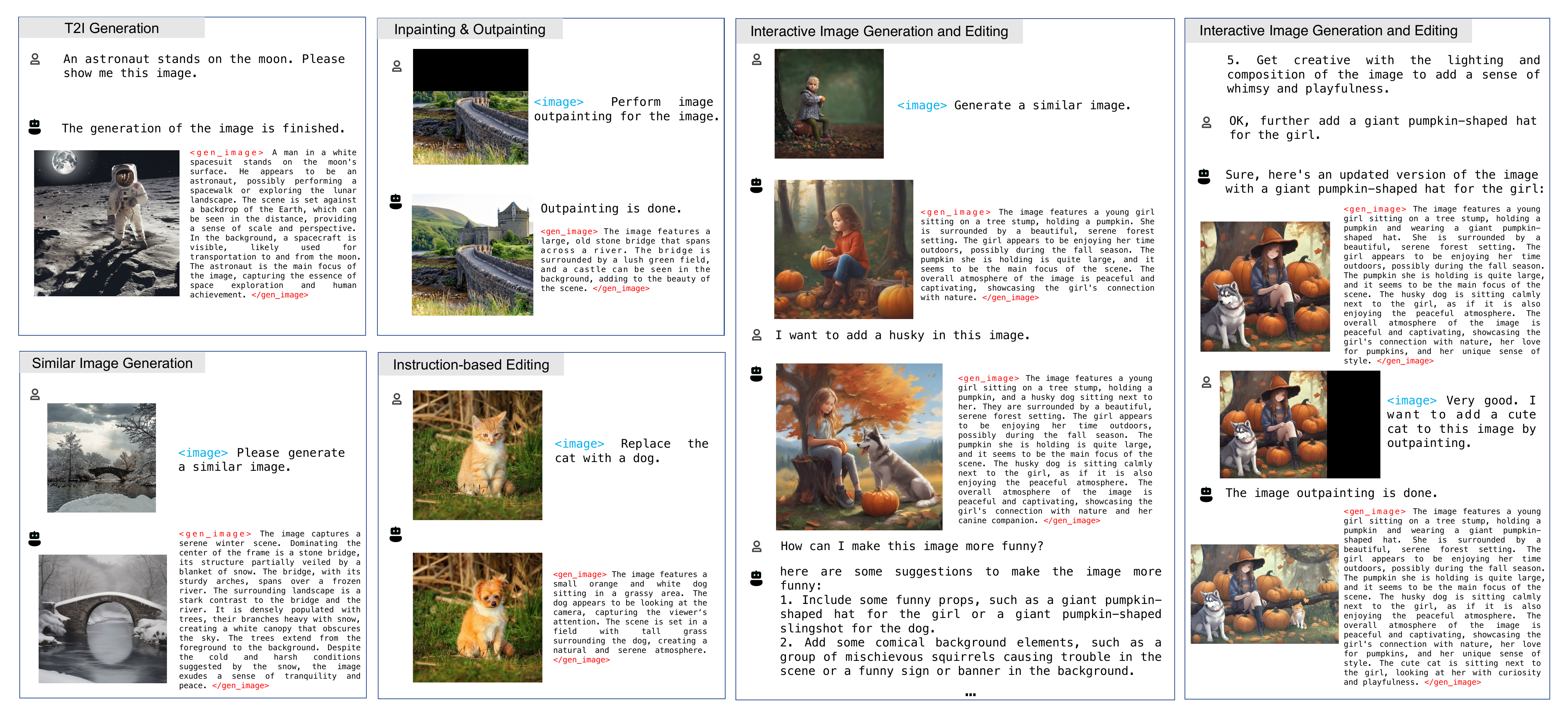}
 \vspace{-6mm}
	\caption{ 
Some examples of LLMGA for assisting in image generation and editing.
\textbf{(1)} T2I generation. LLMGA can refine the user's generation prompt to produce more vivid and vibrant images.
\textbf{(2)} Similar image generation.  LLMGA can understand the component and layout of the input images and generate a similar image.
\textbf{(3)} Inpainting \& Outpainting.  LLMGA can provide detailed generation prompts based on user preferences and input images.
\textbf{(4)} Instruction based editing. LLMGA can understand user instructions and realize accurate editing. 
\textbf{(5)} Interactive image generation and editing exemplify the comprehensive capabilities of LLMGA. Users can design satisfactory images by engaging in interactions with LLMGA, leveraging its vast knowledge. 
}
\label{fig:poster}
  \vspace{-7mm}
\end{figure*}

However, existing works~\cite{GILL,Emu} merely focus on enabling LLM to output images but do not aim to assist users in generating or editing images to enhance quality. In this paper, we aim to develop a Multimodal Large Language Model-based Generation Assistant (LLMGA) to better assist image generation models, making them more user-friendly and capable of producing high-quality images. 
In contrast to certain methods~\cite{GILL, Emu} that leverage MLLMs to predict fixed-size visual embeddings for implicit SD control, our approach is straightforward. We guide the generation of SD using detailed language prompts from MLLM based on five reasons. 
\textbf{(1)} The embeddings predicted by the MLLM are often filled with noise. This can be filtered out by mapping them to a fixed language domain, enabling precise control of SD.
\textbf{(2)} Detailed language prompts can make the network more transparent and interactive, allowing users to understand MLLM's thoughts for generating images.  
\textbf{(3)}  MLLM is pre-trained on vast textual datasets. Explicit language prompts rather than implicit embeddings are more advantageous for MLLM to generate prompts and comprehend context.
\textbf{(4)} Dynamic-sized language prompt facilitates the addition of generation requests during interactions.
\textbf{(5)} Training is more simple and efficient.

However, we face several challenges:
\textbf{(1)} MLLM may reject the execution of generation instructions due to its nature as a language assistant.
\textbf{(2)} MLLM lacks a comprehensive understanding of image generation and editing, and cannot provide an accurate and detailed generation prompt.
\textbf{(3)} Determining which part of texts generated by MLLM to guide SD generation.
\textbf{(4)} SD's CLIP encode only $75$ tokens. Additionally, SD is trained on short captions, whereas our LLMGA typically generates detailed prompts exceeding $150$ tokens. This discrepancy poses a challenge for SD in following the detailed prompt of LLMGA.

To this end, we have devised a two-stage training scheme. First, we construct a training dataset: prompt refinement, similar image generation, inpainting \& outpainting, and visual question answering. 
We then train LLMGA on these four datasets to cultivate four fundamental capabilities: \textbf{(1)}  For concise user prompts, LLMGA can refine the generation of intricate details, encompassing attire, background, and characters. \textbf{(2)} LLMGA can precisely regenerate an image it observes. \textbf{(3)} LLMGA can generate or refine prompts for inpainting \& outpainting based on its understanding of the image. \textbf{(4)} LLMGA can generate accurate prompts for instruction-based editing according to users' requirements and given images. Additionally, we make LLMGA use special symbols  \texttt{<gen\_img>} and \texttt{</gen\_img>} to distinguish generation prompts and responses.
In the second stage, we freeze the parameters of LLMGA's MLLM and initiate joint training with the SD. This process enables the SD to acclimate to the detailed prompt produced by the MLLM. Notably, when the input token count exceeds $75$, we iteratively apply the CLIP~\cite{CLIP} encoder to the surplus tokens.

Moreover, we have identified noticeable disparities in texture, contrast, and brightness between the newly generated and preserved sections in inpainting \& outpainting. Therefore, we propose a Diffusion-based Reference Restoration Network (DiffRIR). Specifically, aside from images generated by SD, we add masked images as reference inputs into DiffRIR. This enables the DiffRIR to refer to the texture, contrast, and brightness of the preserved regions for restoration. Additionally, we introduce perturbations to contrast and brightness during training, enabling DiffRIR to correct contrast and brightness disparities in the images.

As shown in Fig.~\ref{fig:poster}, LLMGA is a unified and interactive framework for image generation and editing, endowed with a wide array of capabilities: \textbf{(1)} LLMGA leverages its extensive world knowledge and powerful reasoning abilities to assist image generation and editing and significantly improve results. \textbf{(2)} LLMGA can be integrated with external plugins, like ControlNet. \textbf{(3)} Most importantly, users can interact with LLMGA to design satisfying images in a more convenient, flexible, and enjoyable way. In summary, our contributions are as follows:
 \begin{itemize}
 \vspace{-1mm}
 \item  We proposed LLMGA, a simple yet powerful interactive generation and editing framework. Experiments affirm the efficacy of LLMGA in enhancing generation and editing thanks to its vast knowledge and interactive features.  Plus, LLMGA can integrate with external plugins for wider applications.

 \item We construct a training dataset, including four parts: prompt refinement, similar image generation, inpainting \&~outpainting, and instruction-based editing. This enhances LLMGA's comprehension of generation and editing tasks while standardizing response formats.

 \item  We proposed a restoration network DiffRIR, which introduces reference images and training perturbations to contrast and brightness. DiffRIR can alleviate texture, contrast, and brightness discrepancies between newly generated and preserved regions for edited images.

 \item Open-source. The following assets are released: the generated data, the codebase for model training, the model checkpoint, and a demo.
\end{itemize}

\section{Related Work}
\label{sec:relatework}

\noindent
\textbf{Diffusion Model.}  Diffusion Models (DMs)~\cite{diffusion,DDPM2,song2020score,kingma2021variational,DDPM3,DDPM4,DDPM5,DDPM6,sdedit,peebles2023scalable,bao2023all,instructp2p} have achieved remarkable results in image generation.  DMs adopt a parameterized Markov chain to optimize the lower variational bound on the likelihood function. In this way, it can generate realistic images from Gaussian noise. After that, several DM methods~\cite{diffir,diffi2i, repaint,Palette,bao2023one,fan2023frido,kawar2023imagic,kim2022diffusionclip,valevski2022unitune,batzolis2021conditional,podell2023sdxl,xue2023raphael,feng2023ernie,balaji2022ediffi,saharia2022photorealistic,dai2023emu} have been tailored to enhance the text-to-image (T2I) generation and editing. Notably, GLIDE~\cite{glide} pioneered the incorporation of text features into transformer blocks during the denoising process. Subsequently, DALL-E~\cite{Dalle}, Imagen~\cite{Imagen}, and Stable Diffusion~\cite{LDM} have made substantial strides in improving T2I generation. Subsequently, some works ~\cite{controlnet,dreambooth,sur-adapter} introduced conditioning controls to the DMs to facilitate a more convenient and precise manipulation of the generation process. Overall, enhancing the user-friendliness of DMs is a key focus within the community. In this paper, we introduce LLMGA, leveraging the extensive knowledge and powerful reasoning capabilities of LLM to facilitate users in achieving more easily attainable and satisfactory image designs.

\noindent
\textbf{Multimodal Large Language Models.}
Recently, LLMs have undeniably made profound impacts and revolutions within the entire AI community and beyond. 
For example, exemplary LLMs, such as ChatGPT and GPT4~\cite{gpt4}, have showcased remarkable abilities in comprehension, reasoning, responses, and knowledge reservoirs. 
Subsequently, a range of LLMs~\cite{chung2022scaling}, including Vicuna~\cite{vicuna}, LLaMA~\cite{llama}, and Alpaca~\cite{alpaca} have been released as open-source models, substantially propelling advancements of the community.

Afterward, the community began focusing on the development of the Multimodal Large Language Model~\cite{minigpt4,huang2023language,palm-e,ye2023mplug,dreamllm,seed,cm3leon}. They aim to enable LLMs to comprehend both images and text and provide textual responses. For instance, Flamingo~\cite{flamingo} encodes images and feeds them into the LLM's attention layer. BLIP-2~\cite{BLIP-2} employs Q-Former to encode input images into queries. 
Additionally, LLaVA~\cite{LLAva} leverages CLIP~\cite{CLIP} to encode images into visual embeddings, which are then concatenated with text embeddings.

Recent concurrent works, such as Next-GPT~\cite{next-gpt}, have extended the capabilities to encompass audio and video modalities.
Moreover, Visual-ChatGPT~\cite{visualchatgpt} and HuggingGPT~\cite{hugginggpt} make LLMs act as agents capable of invoking various pre-trained visual models to achieve MLLM. However, these works focus on making LLM determine the combined invocation of modules (such as detection, recognition, and generation) to fulfill user requirements. However, these methods are not tailored for generation and editing and use concise prompts that lack the capability to enhance results. Thus, we propose LLMGA, which is designed to assist with various image generation and editing tasks. It can achieve satisfactory results by strong reasoning capability and flexible interaction with users.

\section{Methodology}

\begin{figure*}[t]
	\centering
 \resizebox{1\linewidth}{!}{
	\includegraphics[height=8cm]{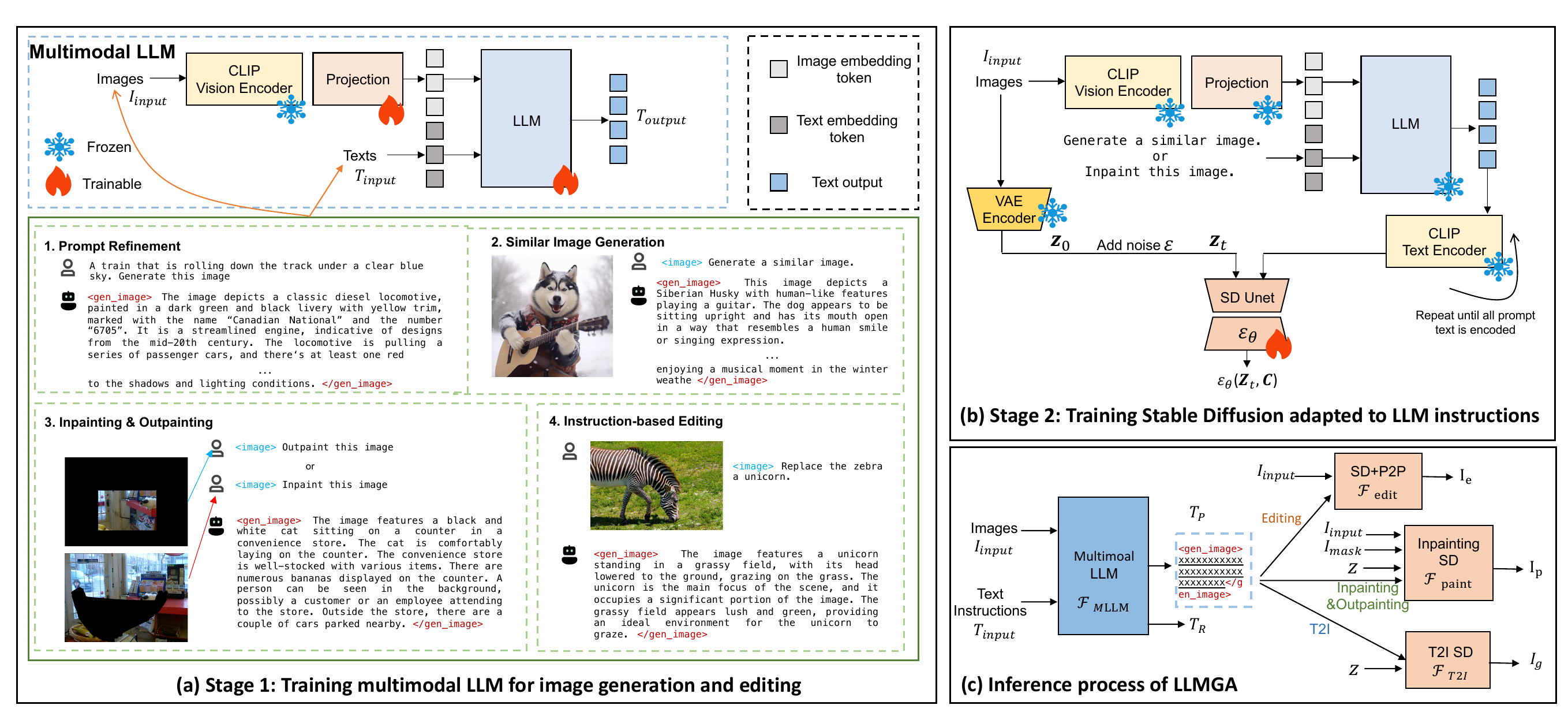}
 }
 \vspace{-6mm}
	\caption{The overview of  LLMGA.
\textbf{(a)} In the first training stage, we train the MLLM to produce generation prompts based on provided instructions. Moreover, we construct a training dataset including four categories: prompt refinement, similar image generation, inpainting \& outpainting, and instruction-based editing.
\textbf{(b)} In the second training stage, we optimize SD to adapt to the detailed generation prompts from MLLM.
\textbf{(c)} In the inference stage, LLMGA can respond to user queries and assist in various tasks, such as image generation, inpainting \& outpainting, and editing.}

\label{fig:method}
 \vspace{-4mm}
\end{figure*}

\subsection{Overview of LLMGA}
\label{sec:overview}

In this paper, we aim to design a MLLM-based Generation Assistant (LLMGA). Our LLMGA produces detailed language-based generation prompts to control SD rather than predicting fixed-sized visual embeddings~\cite{GILL, Emu} to govern SD.
This has five advantages:
\textbf{(1)} Visual embeddings contain noise, and mapping them to the language domain can filter out this noise, enabling precise SD control.
\textbf{(2)} Language-based generation prompts facilitate users in understanding the LLMGA's thoughts, enhancing interaction.
\textbf{(3)} Dynamic-sized language-based generation prompt enables the addition of generation requests.
\textbf{(4)} MLLM is pre-trained on textual datasets. Language prompts rather than implicit visual embeddings are more advantageous for MLLM to generate accurate prompts and comprehend context. 
\textbf{(5)} Training is simpler and more efficient.

However, we need to address several issues:
\textbf{(1)} As a language assistant, MLLM may decline the execution of generation instructions.
\textbf{(2)} MLLM lacks a nuanced understanding of image generation and editing, and cannot produce precise and detailed generation prompts.
\textbf{(3)} MLLM needs to decide which part of the output text serves as generation prompts to guide generation.
\textbf{(4)} SD's CLIP encode only $75$ tokens. Moreover, SD is primarily trained on short captions, while detailed prompts generated by LLMGA may exceed $150$ tokens. This disparity makes it hard for SD to understand the instructions from LLMGA.
To address the aforementioned challenges, we construct a training dataset and design two-stage training schemes to train the MLLM (Sec.~\ref{sec:mllm_tuning}) and SD (Sec.~\ref{sec:sd_tuning}).

The network structure and pipeline of LLMGA are illustrated in Fig.~\ref{fig:method}.  Specifically, as shown in Fig.~\ref{fig:method}~(a)~and~(c), the images $\mathbf{I}_{input}$ are encoded into image embeddings by CLIP vision encoder and a projection layer. Subsequently, the image embedding is concatenated with the text embedding and fed into the LLM to obtain text output $\mathbf{T}_{output}$. This process can be formulated as:
\vspace{-2mm}
\begin{equation}
\label{eq:MLLM}
\vspace{-1mm}
\mathbf{T}_{output}=\mathcal{F}_{MLLM}(\mathbf{T}_{input},\mathbf{I}_{input}),
\end{equation}
where $\mathbf{T}_{input}$ indicates the input text instructions from users. It is notable that  $\mathcal{F}_{MLLM}$ can process only $\mathbf{T}_{input}$ as input.

The text output $\mathbf{T}_{output}$ can comprise two components: text response $\mathbf{T}_{R}$ and generation prompt $\mathbf{T}_{P}$. To distinguish between $\mathbf{T}_{R}$ and $\mathbf{T}_{P}$, we adopt new special tokens, \ie, $\texttt{<gen\_img>}$ and $\texttt{</gen\_img>}$, to encompass $\mathbf{T}_{P}$.

We present $\mathbf{T}_{R}$ as the immediate text response to users. Concurrently, $\mathbf{T}_{P}$ is further fed into the subsequent SD to guide T2I generation (Eq.~\ref{eq:sd_t2i}), inpainting \& outpainting (Eq.~\ref{eq:sd_painting}), and instruction-based image editing (Eq.~\ref{eq:sd_edit}). 
\vspace{-1mm}
\begin{equation}
\label{eq:sd_t2i}
\vspace{-1mm}
\mathbf{I}_{g}=\mathcal{F}_{T2I}(\mathbf{T}_{P},\mathbf{Z}),
\end{equation}
\vspace{-4mm}
\begin{equation}
\label{eq:sd_painting}
\vspace{-1mm}
\mathbf{I}_{p}=\mathcal{F}_{Paint}(\mathbf{I}_{input},\mathbf{I}_{mask},\mathbf{T}_{P},\mathbf{Z}),
\end{equation}
\vspace{-3mm}
\begin{equation}
\label{eq:sd_edit}
\mathbf{I}_{e}=\mathcal{F}_{Edit}(\mathbf{I}_{input},\mathbf{T}_{P}),
\end{equation}
Where $\mathbf{Z}$ denotes the random Gaussian noise. Furthermore, to ensure the encoding of all $\mathbf{T}_{P}$ for SD, we iteratively run the CLIP text encoder until all prompts are encoded.  
For inpainting \& outpainting, except the input image $\mathbf{I}_{input}$, an additional mask $\mathbf{I}_{mask}$ are essential inputs for the inpainting SD to specify the region requiring generation. For instruction-based image editing, $\mathcal{F}_{Edit}$ performs inversion~\cite{DDIM,directinversion} and prompt-to-prompt~\cite{prompt2prompt} based on T2I SD.   

\vspace{-1mm}
\subsection{MLLM Training}
\label{sec:mllm_tuning}
As described in Sec.~\ref{sec:overview}, original MLLMs are specifically designed and trained as language assistants, but they lack the proficiency to assist in image generation,  and editing. 
Notably, considering the input of images and the performance of open-sourced MLLMs, employing few-shot learning to guide the model to achieve the desired results proves challenging and inefficient.
Therefore, it is crucial to train the MLLM to serve as proficient assistants in image generation and editing tasks, understanding the expected response formats and enhancing their comprehension of image generation and editing properties.
To this end, as depicted in Fig.~\ref{fig:method}~(a), we construct a training dataset consisting of four categories:
 \begin{itemize}
 \item  \textit{Prompt Refinement.} 
We establish this dataset to cultivate the prompt refinement ability of the MLLM. Specifically, we utilize GPT4-V to furnish detailed descriptions of images in MSCOCO~\cite{mscoco}. These detailed descriptions, along with the original MSCOCO brief descriptions, constitute a training text pair. During training, we input the brief MSCOCO captions and randomly select and append a generation instruction, such as \textit{``Generate this image''}, or a description instruction, like \textit{"Describe this sentence in detail''}. When a generation instruction is included in the prompt, we add $\texttt{<gen\_img>}$ and $\texttt{</gen\_img>}$ on the later detailed description for training.

\item  \textit{Similar Image Generation.} 
 We select images from the MSCOCO dataset along with corresponding detailed descriptions generated by GPT4-V to create the Similar Image Generation dataset. During training, we input the images along with a generation instruction, such as \textit{``Generate a similar image.''}. In cases where a generation instruction is provided,  we add $\texttt{<gen\_img>}$ and $\texttt{</gen\_img>}$ on the subsequent detailed description for training.

 \item \textit{Inpainting $\&$ Outpainting.} 
We use pairs of detailed descriptions and images from the Similar Image Generation dataset. During training, we input masked images with inpainting or outpainting instructions, like \textit{``Inpaint this image.''} or \textit{``Outpaint this image.''}.  Besides, we include $\texttt{<gen\_img>}$ and $\texttt{</gen\_img>}$ on the subsequent detailed description during training.

 \item  \textit{Instruction-based editing.} 
We fine-tune Mixtral-8x7B~\cite{mixtral} to enable it to automatically generate editing data based on detailed descriptions from MSCOCO. For example, given the original caption ``A cat is lying on the ground'', Mixtral-8x7B generates the output ``{Instruction: Replace the cat with a dog., Target caption: A dog is lying on the ground}''. Subsequently, we clean the generated data. During training, we can input MSCOCO images or original captions and provide corresponding editing instructions, aiming to train LLMGA to output the target caption. Additionally, we include $\texttt{<gen\_img>}$ and $\texttt{</gen\_img>}$ tags on the target caption during training.

\end{itemize}


During training, alongside the above data, we integrated the image designing and Alpaca~\cite{alpaca} dataset, to enhance LLMGA's question-answering (QA). Notably, we excluded certain visual multimodal incompatible question-answer pairs from it. Furthermore, we incorporated the LLaVA v1.5 mix665k dataset~\cite{llava1-5} to endow LLMGA with Visual Question Answering (VQA) capabilities. We provide more details in the supplementary material.

As illustrated in Fig.~\ref{fig:method}~(a), we freeze the CLIP vision encoder and optimize the projection layer and LLM. The model is trained end-to-end using the auto-regressive cross-entropy loss ($\mathcal{L}_{MLLM}$) for text generation. Given the ground-truth targets $\mathbf{T}_{GT}$, this loss can be formulated as: 
\vspace{-1mm}
\begin{equation}
\label{eq:loss_mllm_sup}
\mathcal{L}_{MLLM}=\mathbf{C E}\left(\mathbf{T}_{output}, \mathbf{T}_{GT}\right).
\end{equation}

\vspace{-3mm}
\subsection{Stable Diffusion Training}
\label{sec:sd_tuning}

As described in Sec.~\ref{sec:overview}, the original SD's CLIP text encoder only encodes $75$ tokens, which cannot handle the entire MLLM's generation prompt. Moreover, the original SD is trained on brief captions, which cannot fully understand the generation prompts. 
Thus, we repeatedly use the CLIP text encoder to encode all instances of $\mathbf{T}_{P}$ for SD. Besides, we train the T2I SD model and the inpainting SD model, respectively. 
For both generation and inpainting \& outpainting tasks, the generation prompts of MLLM are detailed descriptions of images.  Therefore, during training, we can instruct MLLM to provide a detailed description $\mathbf{T}_{P}$ for images from the LAION-Aesthetics~\cite{laion5b} and MSCOCO datasets. Subsequently,  $\mathbf{T}_{P}$ is fed into T2I SD or inpainting SD for joint training.
Notably, we only optimize the SD unet while freezing the parameters of other networks. To accelerate the training, we record the prompt $\mathbf{T}_P$ of MLLM to avoid redundant calculations. The model is trained using SD loss (Eq.~\ref{eq:sd_loss}). For instruction-based editing, we directly adopt the pre-trained T2I SD model.
\vspace{-2mm}
\begin{equation}
\vspace{-2mm}
\mathcal{L}_{SD}=\mathbb{E}_{\mathbf{Z}_t, \mathbf{C}, \epsilon, t}\left(\left\|\epsilon-\epsilon_\theta\left(\mathbf{Z}_t, \mathbf{C}\right)\right\|_2^2\right),
\label{eq:sd_loss}
\end{equation}
where $\mathbf{Z}_t=\sqrt{\overline{\alpha_t}} \mathbf{Z}_0+\sqrt{1-\overline{\alpha_t}} \epsilon$ represents
the noised feature map at timestep $t$. Ground truth images are encoded into latent space to derive $\mathbf{Z}_0$. Here, $\epsilon \in \mathcal{N}(0, \mathbf{I})$ represents Gaussian noise, and $\epsilon_\theta$ refers to the SD unet. $\mathbf{C}$ indicates the conditional information. For T2I generation, $\mathbf{C}$ is $\mathbf{T}_{P}$. For inpainting and outpainting,  $\mathbf{C}$ contains $\mathbf{T}_{P}$, the mask, and the VAE-encoded masked image.

\vspace{-1mm}
\subsection{Restoration Network Training}
\label{sec:restoration}

For SD inpainting \& outpainting, we observed noticeable disparities between the preserved and newly generated regions in the edited images. These disparities can be attributed to variations in texture, contrast, and brightness. Besides, after finishing image generation and editing, it is customary to employ image restoration methods~\cite{diffir,diffi2i,real-esrgan,KDSR} on the generated images to enhance the quality. Therefore, we intend to employ an image restoration network to address these disparities in the inpainting \& outpainting images.

To enhance the consistency between the newly generated and the preserved regions, we introduce a reference-based restoration scheme. Specifically, existing restoration methods~\cite{diffir,real-esrgan,KDSR} take low-quality (LQ) images as input and produce high-quality (HQ) images, but they often do not leverage the preserved information from the given masked image. Different from them, we concatenate the LQ image ${I}_{LQ}$ and the masked image, \ie, $(1-\mathbf{I}_{mask})\mathbf{I}_{GT}$, as inputs and feed them into the restoration model $\mathcal{F}_R$. This process can be formulated as:
\vspace{-2mm}
\begin{equation}
\vspace{-2mm}
\mathbf{I}_{HQ}=\mathcal{F}_R(concat(\mathbf{I}_{LQ},(1-\mathbf{I}_{mask})\mathbf{I}_{GT})).
\end{equation}

To further mitigate the brightness and contrast disparities, we introduced additional color degradation (\ie, random  brightness and contrast disturbance) into the training process of the restoration model, which is formulated as:
\vspace{-2mm}
\begin{equation}
\vspace{-5mm}
\mathcal{D}_2(\mathbf{x})=c_1\mathbf{x}+c_2,
\end{equation} 
\begin{equation}
\vspace{-1mm}
\mathbf{I}_{LQ}=\mathcal{D}_2(\mathcal{D}_1(\mathbf{I}_{GT})),
\end{equation} 
where $\mathcal{D}_2$ indicates contrast and brightness disturbance, $\mathbf{x}$ denotes the input image. $c_1$ is the contrast gain, randomly varied within the range of $[0.94, 1.06]$, while $c_2$ is the brightness bias, randomly varied within the range of $[-0.05, 0.05]$.
$\mathcal{D}_1$ represents the real-world degradation process used in restoration model~\cite{real-esrgan,diffir,KDSR}. $\mathbf{I}_{GT}$ is ground truth image, and $\mathbf{I}_{LQ}$ is LQ image.
Here, we adopt the network structure of the SOTA restoration model DiffIR~\cite{diffir} and apply our schemes to it to obtain our Diffusion-based Reference Restoration Network (DiffRIR). 

\vspace{-3mm}
\section{Experiments}
\vspace{-1mm}

\subsection{Implementation Details}
For the first stage of training, we employ the pretrained LLaVA-1.5-7B or LLaVA-1.5-13B as the initial MLLM. We utilize the AdamW optimizer, setting the learning rate to $2\times10^{-5}$. Moreover, we adopt CosineLR as the learning rate scheduler. The batch size per device and epochs are set to $16$ and $1$, respectively. Besides, the training ratios for VQA (LLaVA v1.5 mix665k), QA (Alpaca), prompt refinement, similar image generation, inpainting \& outpainting, and instruction-based editing are specified as $1$, $0.3$, $0.3$, $0.3$, $0.3$ and $0.3$, respectively.

For the second stage of training, we adopt the Stable Diffusion 1.5 (SD1.5) as the initial image generation or inpainting \& outpainting model. We train these models with the AdamW optimizer, setting the learning rate to $1\times10^{-5}$. The batch size is set to $32$. We train SD1.5 by $1\times10^{5}$ iterations.

For the restoration network, we train DiffRIR on DIV2K~\cite{DIV2K} and Flickr2K~\cite{Flickr2K} datasets with the same GAN-based loss function as DiffIR. The batch sizes are set to $64$, and the
LQ patch sizes are set to $64\times64$. We use Adam optimizer, setting the learning rate to $2\times10^{-4}$. We train this model by $4\times10^{5}$ iterations.

\begin{wraptable}{r}{0.5\linewidth}
\centering
\vspace{-12mm}
\centering
\caption{Quantitative comparison on \textbf{T2I} generation on the  MSCOCO~\cite{mscoco} dataset.}
\resizebox{1\linewidth}{!}{
    \begin{tabular}{lcc}
    \toprule[0.2em]
    \textbf{Method} &  \multicolumn{1}{c}{\textbf{FID}$\downarrow$} & \multicolumn{1}{c}{\textbf{IS}$\uparrow$} \\
    \midrule
    SD1.5 & 24.0081  & 35.99  \\
    \midrule
    GILL~\cite{GILL}  & 25.1123  & 34.20  \\
    \midrule
    LLMGA-7b(SD1.5) & 23.5946  & 37.12  \\
    LLMGA-7b(SD1.5-ft, Ours) & \underline{18.5234}  & \underline{41.04}  \\
    \midrule
    LLMGA-13b(SD1.5) & 23.5828  & 37.58  \\
    LLMGA-13b(SD1.5-ft, Ours) & \textbf{18.4063}  & \textbf{41.16}  \\
    \bottomrule[0.2em]
    \end{tabular}%
    }
  \label{tab:T2I}
\vspace{-7mm}
\end{wraptable}

\vspace{-1mm}
\subsection{Experimental Results}

\textbf{Evaluation on T2I Generation.} The results are shown in Tab.~\ref{tab:T2I}. Notably, SD1.5 is the original Stable Diffusion 1.5 while SD1.5-ft indicates the finetuned SD1.5 in LLMGA. We also compare our LLMGA with the recently proposed multimodal generative model GILL~\cite{GILL}. \textbf{(1)} Comparing SD1.5, our LLMGA-7B achieves notable 5.4847 FID and 5.05 IS improvements, underscoring the effectiveness of LLMGA. \textbf{(2)} Moreover, our LLMGA-7B significantly outperforms GILL. \textbf{(3)} In the 3rd and 4th rows of Tab.~\ref{tab:T2I}, we use LLMGA-7B to refine the short MSCOCO caption to the detailed generation prompt and send it to the SD1.5 and SD1.5-ft, respectively. Our LLMGA-7B with SD1.5-ft achieves significant 5.0712 FID and 3.92 IS improvements, respectively. This demonstrates our second-stage training can help SD1.5-ft to better follow detailed prompts from MLLM. 
\textbf{(4)} Comparing the 4th and 6th rows of Tab.~\ref{tab:T2I}, LLMGA-13B exhibits better performance than LLMGA-7B due to its superior reasoning ability. 

\begin{figure*}[t]
	\centering
 \resizebox{1\linewidth}{!}{
	\includegraphics[height=4cm]{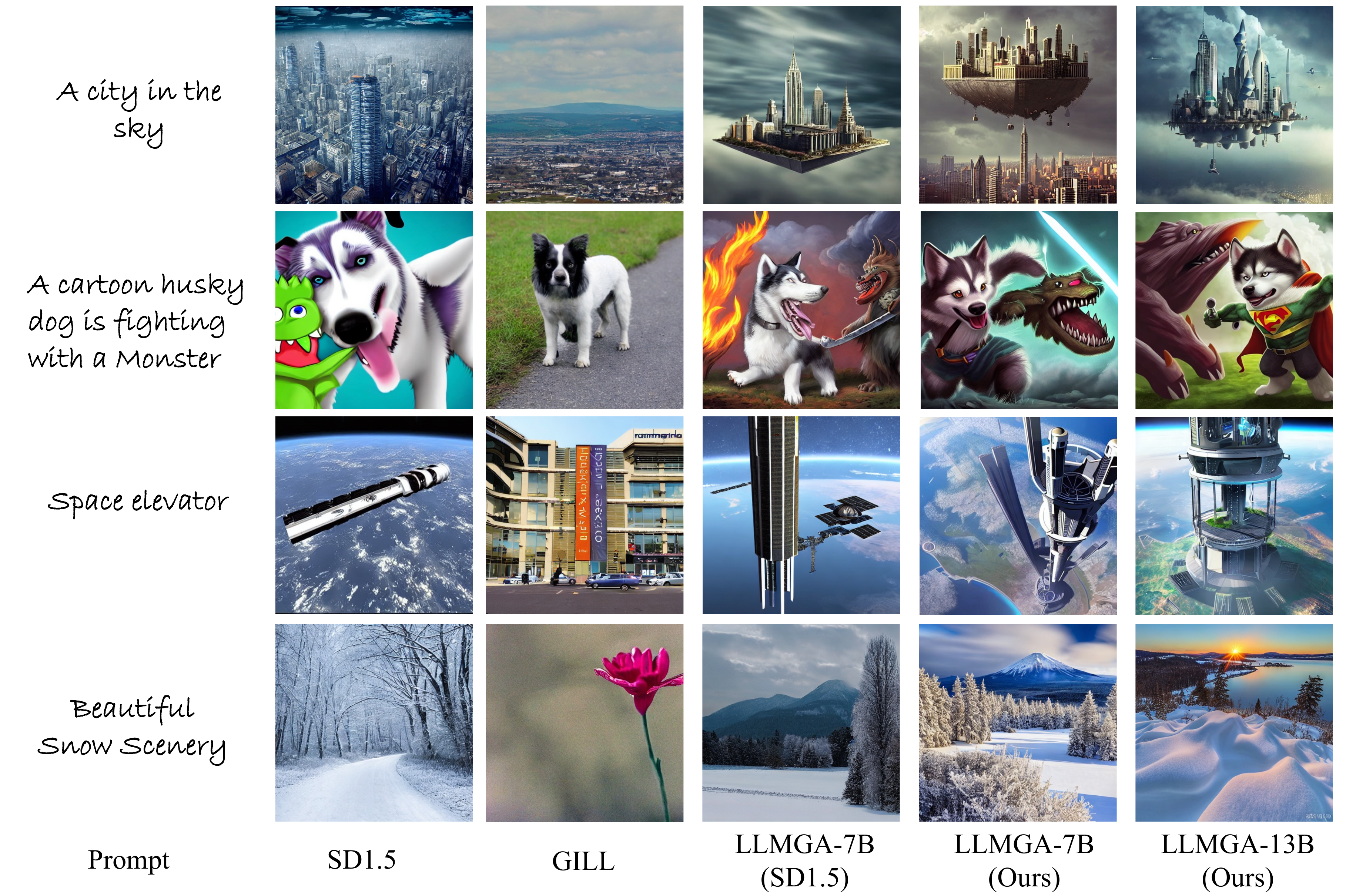}
 }
  \vspace{-6mm}
	\caption{\textbf{T2I} visual comparison. LLMGA can produce accurate and high-quality results.    }
	\label{fig:t2i}
 \vspace{-1mm}
\end{figure*}

The qualitative results are shown in Fig.~\ref{fig:t2i}. \textbf{(1)} LLMGA excels in refining prompts by incorporating details to generate visually rich and pleasing images. For example, in the first row, LLMGA crafts a battle attire for the husky, depicting engaging scenarios of battling monsters. This makes user usage more convenient, eliminating the need for them to think about generating image details themselves.  \textbf{(2)} LLMGA can leverage its extensive knowledge base to generate images, even for concepts users may not be familiar with, like a space elevator.

\begin{table}[t]
  \centering
 
  \caption{Quantitative comparison for \textbf{image editing} on MagicBrush test set~\cite{magicbrush}.}
   \vspace{-2mm}
    \begin{tabular}{c|ccccc}
    \toprule[0.2em]
    \textbf{Methods} & \textbf{L1}$\downarrow$    & \textbf{L2}$\downarrow$    & \textbf{CLIP-I}$\uparrow$ & \textbf{DINO}$\uparrow$  & \textbf{CLIP-T}$\uparrow$ \\
    \midrule
    InstructPix2Pix & 0.1197 & 0.0416 & 0.8442 & 0.7252 & 0.2909 \\
    MagicBrush & \textbf{0.0647} & \underline{0.0224} & \textbf{0.9293} & \textbf{0.8913} & \underline{0.2979} \\
    \midrule
    LLMGA (Ours) & \underline{0.0814} & \textbf{0.0218} & \underline{0.8936} & \underline{0.8768} & \textbf{0.3137} \\
    \bottomrule[0.2em]
    \end{tabular}%
    \vspace{-5mm}
  \label{tab:edit}%
\end{table}%

\begin{figure*}[t]
	\centering
 \resizebox{1\linewidth}{!}{
	\includegraphics[height=4cm]{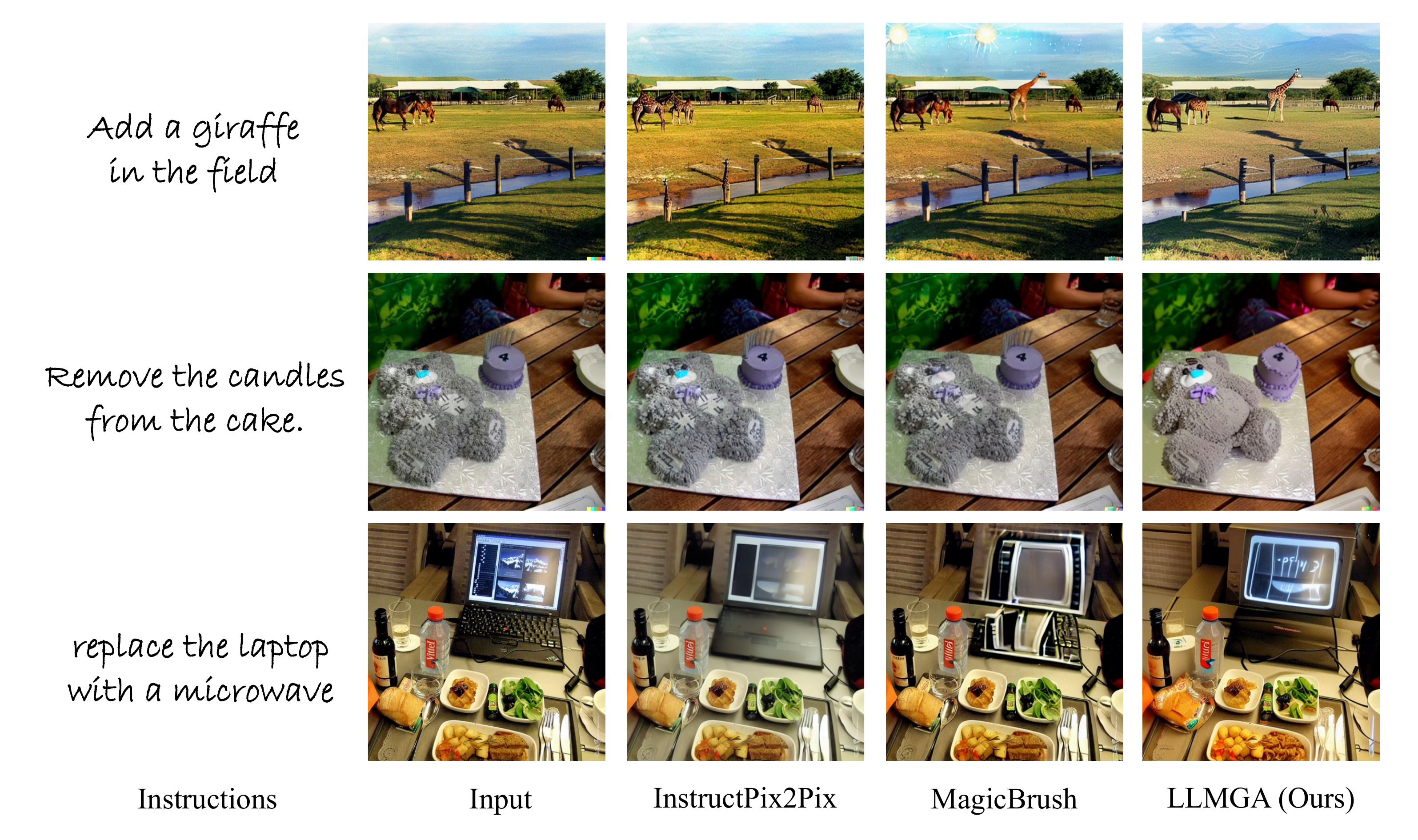}
 }
  \vspace{-6mm}
	\caption{Visual comparison on \textbf{instruction-based editing}.   }
	\label{fig:editing}
 \vspace{-3mm}
\end{figure*}

\noindent\textbf{Evaluation on Instruction-based Editing.} 
For instruction-based editing, we utilize LLMGA to provide a detailed description of the edited image based on the input image and user editing instructions. We then employ Direct Inversion~\cite{directinversion} and prompt-to-prompt~\cite{prompt2prompt} methods to obtain the edited image. 
\textbf{(1)} The quantitative results are presented in Tab.~\ref{tab:edit}. Notably, our LLMGA did not train SD for image editing like InstructPix2Pix~\cite{instructp2p} and MagicBrush~\cite{magicbrush}. However, our zero-shot performance on the MagicBrush test set surpassed that of InstructPix2Pix, achieving performance similar to that of MagicBrush.
\textbf{(2)} Additionally, our LLMGA offers a superior user experience in interactive editing, allowing image modifications to be carried out conversationally. In contrast, InstructPix2Pix only supports input via instructions and output as images. Visualized results are shown in Fig.~\ref{fig:editing}, by leveraging the powerful reasoning capabilities of LLM, our LLMGA can provide more accurate and reasonable editing results.

\begin{table}[t]
  \centering
  \caption{Quantitative comparison for \textbf{outpainting \& inpainting} on Places~\cite{places2}. }
  \vspace{-2mm}
  \resizebox{1\linewidth}{!}{
    \begin{tabular}{c|cccc|cccc}
    \toprule[0.2em]
    \multirow{3}[6]{*}{\textbf{Method}} & \multicolumn{4}{c|}{\textbf{Outpainting}} & \multicolumn{4}{c}{\textbf{Inpainting}} \\
\cmidrule{2-9}          & \multicolumn{2}{c}{\textbf{Narrow Masks}} & \multicolumn{2}{c|}{\textbf{Wide Masks}} & \multicolumn{2}{c}{\textbf{Narrow Masks}} & \multicolumn{2}{c}{\textbf{Wide Masks}} \\
\cmidrule{2-9}          & \textbf{FID}$\downarrow$   & \textbf{LPIPS}$\downarrow$ & \textbf{FID}$\downarrow$   & \textbf{LPIPS}$\downarrow$ & \textbf{FID}$\downarrow$   & \textbf{LPIPS}$\downarrow$ & \textbf{FID}$\downarrow$   & \textbf{LPIPS}$\downarrow$ \\
    \midrule
    SD1.5 & 2.0167  & 0.2283  & 5.0090  & 0.3734  & 1.0795  & 0.1236  & 1.2855  & 0.1434  \\
    \midrule
    LLMGA-7b (SD1.5 ) & 1.5530  & 0.2263  & 3.2630  & 0.3688  & 1.0692    & 0.1233  & 1.0983  & 0.1427  \\
    LLMGA-7b (SD1.5-ft, Ours) & \underline{1.2973}  & \underline{0.2215}  & \underline{2.4409}  & \underline{0.3616}  & \underline{0.8027}   & \underline{0.1171}  & \underline{0.9807}  & \underline{0.1405}  \\
    \midrule
    LLMGA-13b (SD1.5) & 1.5631  & 0.2263  & 3.0845  & 0.3679  & 1.0326   & 0.1228  & 1.0978  & 0.1426  \\
    LLMGA-13b (SD1.5-ft, Ours) & \textbf{1.2160}  & \textbf{0.2210}  & \textbf{2.3663}  & \textbf{0.3609}  & \textbf{0.7992 }  & \textbf{0.1166}  & \textbf{0.9780}  & \textbf{0.1400}  \\
    \bottomrule[0.2em]
    \end{tabular}%
    }
    \vspace{-5mm}
  \label{tab:inpainting}%
\end{table}%

\begin{figure*}[t]
	\centering
 \resizebox{1\linewidth}{!}{
	\includegraphics[height=4cm]{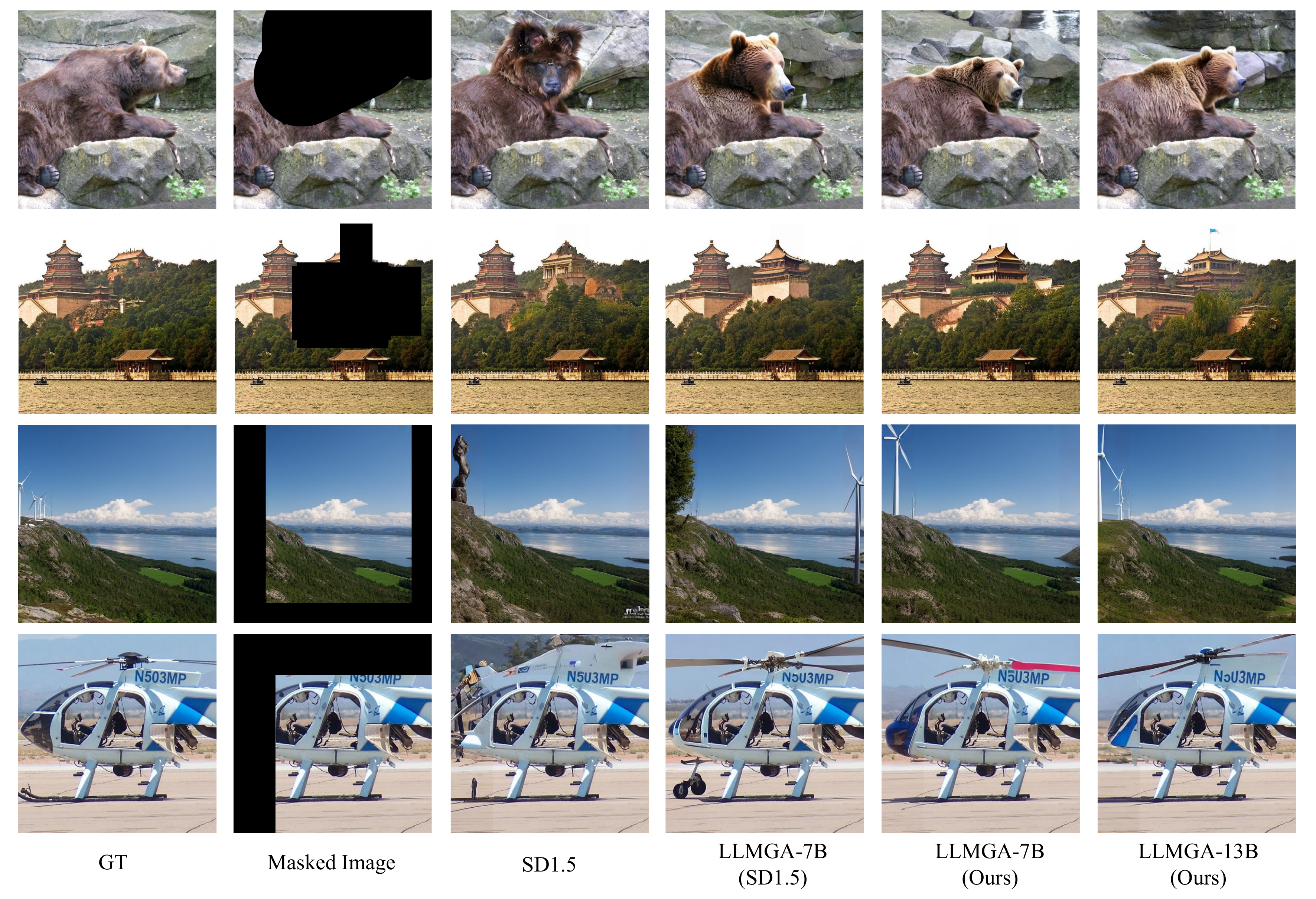}
 }
  \vspace{-6mm}
	\caption{Visual comparison on \textbf{inpainting and outpainting}.   }
	\label{fig:inpainting}
 \vspace{-1mm}
\end{figure*}

\begin{figure*}[th]
\Large
\centering
\resizebox{1\linewidth}{!}{
 \includegraphics[height=4cm]{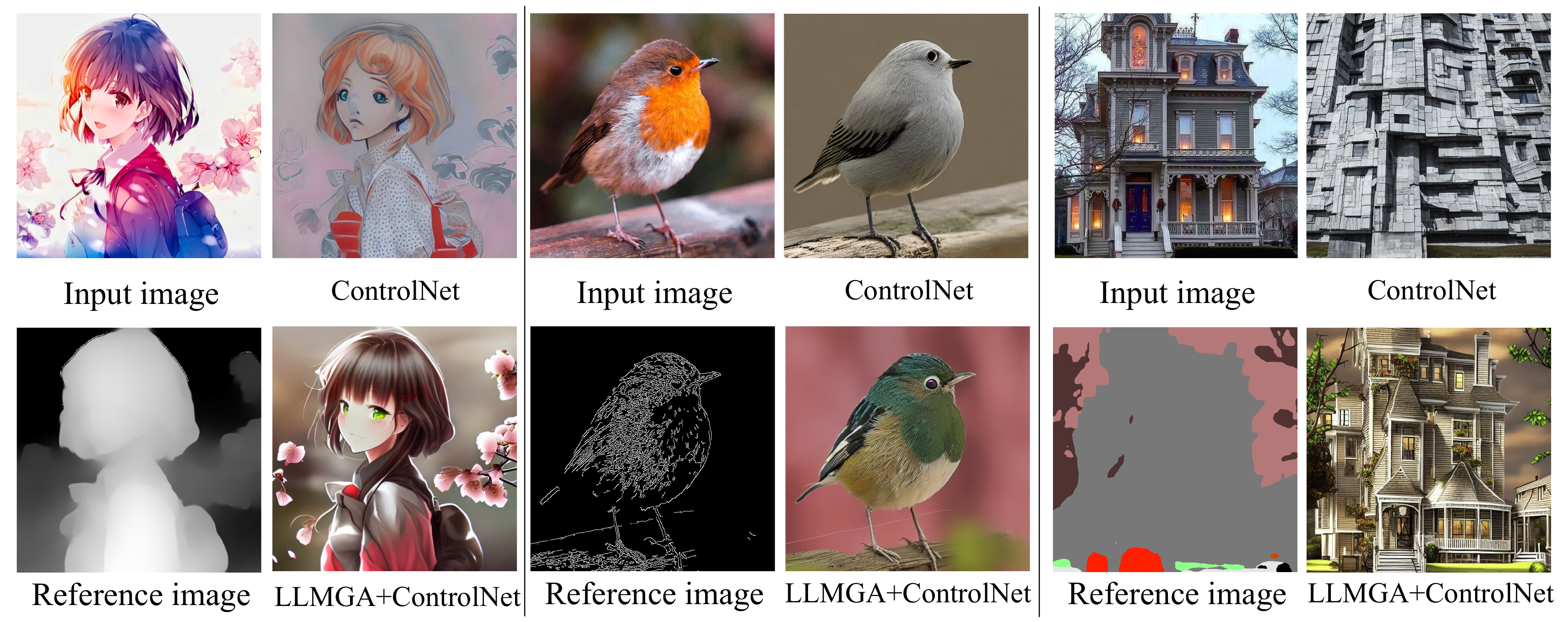 }    
}
\vspace{-9mm}
\caption{Visualization of LLMGA plus ControlNet. Our LLMGA can enhance the details in generated images, producing visually pleasing images.
}
\vspace{-1mm}
\label{fig:control}
\end{figure*}

\begin{table}[t]
  \centering
  \caption{Quantitative comparison on \textbf{image restoration} for   Places~\cite{places2} outpainting.}
  \vspace{-1mm}
  \resizebox{0.82\linewidth}{!}{
    \begin{tabular}{l|cc|cc}
    \toprule[0.2em]
    \textbf{Method} & \textbf{Reference} & \textbf{\shortstack{Color \\Degradation}} & \textbf{FID}$\downarrow$   & \textbf{LPIPS}$\downarrow$ \\
    \midrule
    LLMGA-7B & \XSolidBrush     & \XSolidBrush     &   2.4409    & 0.3616 \\
    LLMGA+DiffIR~\cite{diffir} & \XSolidBrush     & \XSolidBrush     &   2.3587    & 0.3612  \\
    LLMGA+DiffRIR$_1$ & \Checkmark     & \XSolidBrush     &   2.2993     & 0.3609  \\
    \shortstack{LLMGA+DiffRIR$_2$ (Ours)} & \Checkmark     & \Checkmark     &   2.2687     & 0.3607  \\
    \bottomrule[0.2em]
    \end{tabular}%
    }
    \vspace{-4mm}
  \label{tab:restore}%
\end{table}%

\begin{figure*}[th]
\Large
\centering
\resizebox{1\linewidth}{!}{
 \includegraphics[height=4cm]{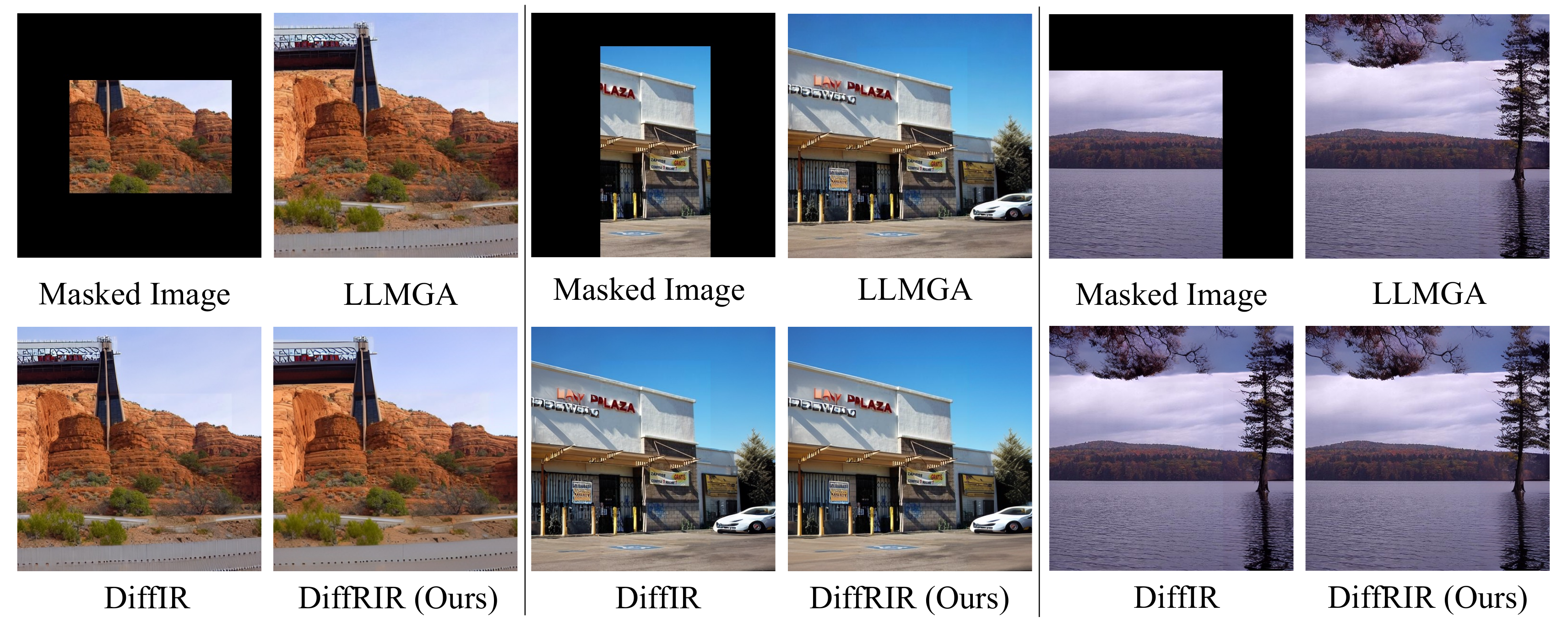 }    
}
\vspace{-9mm}
\caption{Visual comparison of image \textbf{restoration} methods.  DiffRIR can alleviate the texture, contrast, and brightness disparities in inpainting \& outpainting results.  
}
\vspace{-3mm}
\label{fig:IR}
\end{figure*}

\noindent\textbf{Evaluation on Inpainting and Outpainting.} The results are shown in Tab.~\ref{tab:inpainting}. 
\textbf{(1)} Comparing the 1st and 3rd rows of Tab.~\ref{tab:inpainting}, our LLMGA-7B achieves significant improvements of 2.5681 FID over the SD1.5 in outpainting under wide masks. 
\textbf{(2)} In the 2nd and 3rd rows of Tab.~\ref{tab:inpainting}, we make LLMGA imagine the complete generation prompts for the given masked images, which are then input into the later SD. Notably, our LLMGA-7B (with SD1.5-ft) demonstrates a significant FID improvement over LLMGA-7B (with SD1.5) in both outpainting and inpainting. This demonstrates the second stage of training makes SD better follow the prompts from MLLM.
\textbf{(3)} Comparing the 3rd and 5th rows, our LLMGA-13B outperforms LLMGA-7B due to its superior reasoning capabilities.

The qualitative results are shown in Fig.~\ref{fig:inpainting}. We can see that LLMGA can deduce and imagine complete images based on masked input images. For example, in the 3rd row of Fig.~\ref{fig:inpainting}, LLMGA can infer the presence of wind turbines on the mountain based on the given environment. Overall,  LLMGA's powerful reasoning capability and extensive knowledge can assist users in conveniently making accurate and visually pleasing inpainting and outpainting.

\noindent\textbf{Evaluation on ControlNet.} Our LLMGA demonstrates exceptional scalability, enabling integration with external plugins like ControlNet~\cite{controlnet}. Here, we utilize LLMGA to create detailed prompts derived from input images and user requirements, working alongside ControlNet to guide image generation. As depicted in Fig.~\ref{fig:control}, our LLMGA significantly enhances the diversity and richness of outcomes in picture reference-guided image generation. Furthermore, more external plugins can also be integrated into the interactive framework of LLMGA, combining with LLMGA's reasoning design capabilities and all previous functions to allow for a broader range of creative and engaging applications.

\noindent\textbf{Evaluation on Image Restoration.} 
The results are shown in Tab.~\ref{tab:restore}. For comparisons, we validate DiffIR~\cite{diffir} and our DiffRIR on the outpainting image generated by LLMGA. \textbf{(1)} Comparing the 2nd and 3rd rows of Tab.~\ref{tab:restore}, it is evident that introducing a reference scheme can significantly improve restoration performance. \textbf{(2)} When comparing the 3rd and 4th rows of Tab.~\ref{tab:restore}, it can be observed that introducing color degradation helps alleviate the bright and contrast distortion caused by SD. 
\textbf{(3)} Comparing the 1st and 4th rows of Tab.~\ref{tab:restore}, our DiffRIR yields significant improvement, validating the effectiveness of DiffRIR.  

As shown in Fig.~\ref{fig:IR}, our DiffRIR (\ie, DiffRIR$_2$ in Tab.~\ref{tab:restore}) can alleviate the texture, brightness, and contrast discrepancies, and generate realistic details.

\begin{wraptable}{r}{0.5\linewidth}
\centering
\vspace{-2mm}
\centering

\resizebox{1\linewidth}{!}{
\includegraphics[height=4cm]{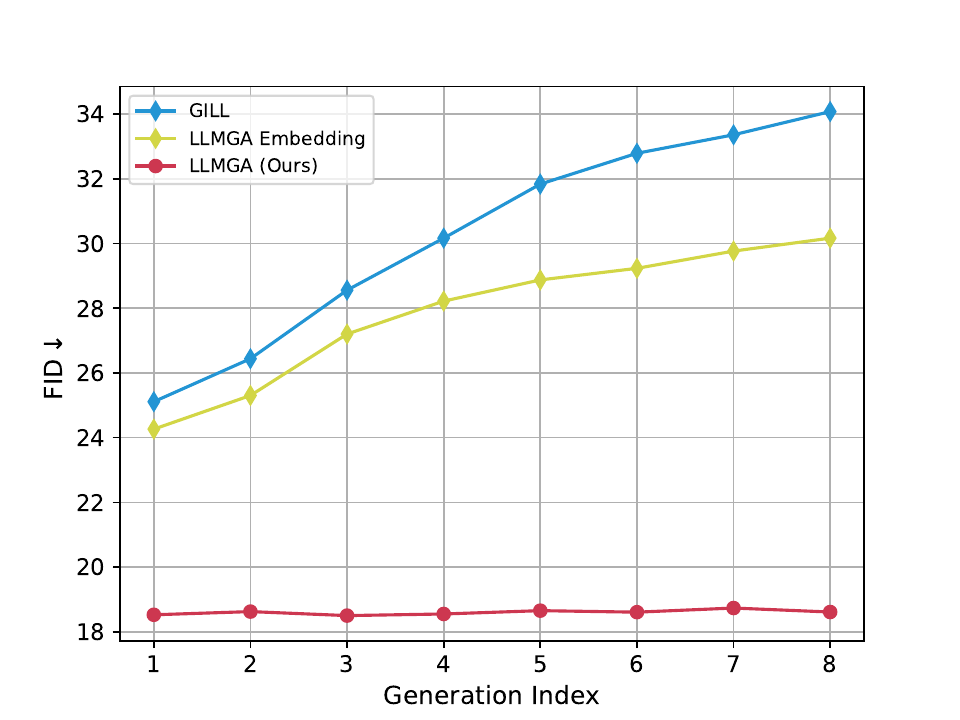}
    }
\vspace{-7mm}
\caption{T2I performance comparison of SD control schemes based on detailed language prompt and embedding. }
\label{fig:poe}
\vspace{-7mm}
\end{wraptable}

\noindent\textbf{Control SD using detailed language prompt or embedding?} The results are shown in Fig.~\ref{fig:poe}.  We compare two approaches: GILL~\cite{GILL}, which makes LLM estimate a fix-sized embedding
to control SD generation, and LLMGA Embedding, a variant of LLMGA where the language prompt is replaced with embedding, undergoing the same training process as LLMGA.  The evaluation is conducted on MSCOCO by instructing these methods to generate images with the same prompts in multiple times in conversation form. \textbf{(1)} The quality of generated images (Fig.~\ref{fig:poe}) in embedding-based methods (\ie, GILL and LLMGA Embedding)  deteriorate rapidly as the number of conversation turns increases. In contrast, our LLMGA remains unaffected. This discrepancy arises from the inherent noise present in the embeddings predicted by LLM.  As the number of conversation turns rises, these generated embeddings integrate with the preceding conversations, introducing even more noise.  This poses challenges for the precise control of SD-generated content. Our LLMGA addresses this issue by mapping the embedding to the fixed language domain, effectively eliminating such noise. \textbf{(2)} Additionally, LLMGA Embedding also outperforms GILL, indicating that the prompt size used to guide SD generation should be adaptive in content, rather than a fixed size.

\begin{table*}[t]
  \centering
  \caption{Datasets comparison. We conducted comparisons of FID on T2I and outpainting. The \Checkmark signifies the utilization of the complete dataset during training. Conversely, the absence of the \Checkmark indicates a reduction to only $10\%$ of the original datasets.}
  \vspace{-3mm}
  \resizebox{1\linewidth}{!}{
    \begin{tabular}{l|cccc|cc}
    \toprule[0.2em]
    \textbf{Method} & \textbf{\shortstack{Prompt \\ Refinement}} & \textbf{\shortstack{Inpainting \\ \&Outpainting}} & \textbf{\shortstack{Similar \\Image Generation}} & \textbf{\shortstack{Instruction-based \\Editing}} & \textbf{T2I}   & \textbf{Outpainting}  \\
    \midrule
    LLMGA$_1$ &       & \Checkmark     & \Checkmark     & \Checkmark     & 21.4460   & 2.6533   \\
    LLMGA$_2$ & \Checkmark     &       & \Checkmark     & \Checkmark     & 19.1914  & 3.1054   \\
    LLMGA$_3$ & \Checkmark     & \Checkmark     &       & \Checkmark     & 19.5698  & 2.6612   \\
    LLMGA$_4$ & \Checkmark     & \Checkmark     & \Checkmark     &       & 19.0642  & 2.8047   \\
    LLMGA$_5$ (Ours) & \Checkmark     & \Checkmark     & \Checkmark     & \Checkmark     & 18.5234 & 2.4409   \\
    \bottomrule[0.2em]
    \end{tabular}%
    }
    \vspace{-5mm}
  \label{tab:dataset}%
\end{table*}%

\noindent\textbf{Contribution of Training Data.}  To assess the impact of training data, we downsized one of the four training datasets in LLMGA$_5$ to $10\%$ of its original magnitude, ensuring that LLMGA remains capable of furnishing responses in the prescribed format. The results are shown in Tab.~\ref{tab:dataset}. It is evident that prompt refinement, inpainting \& outpainting, and instruction-based editing datasets enhance LLMGA's comprehension of image generation and editing properties, resulting in superior images. Moreover, comparing LLMGA$_3$ and LLMGA$_5$, we can see that engaging in similar image generation training further improves the performance of LLMGA in both generation and editing. 

\vspace{-2mm}
\section{Conclusion}
\vspace{-1mm}
LLM possesses an extensive reservoir of knowledge and powerful comprehension and reasoning capabilities. In this paper, we introduce a MLLM-based generation assistant (LLMGA), aiming to exploit LLM's capabilities in an interactive manner to facilitate more efficient and convenient image generation and editing. Compared to relying on LLM to predict a fixed-size embedding to control SD, we employ detailed generation prompts. These prompts prove to be more favorable for enhancing LLM's contextual comprehension and generating more accurate and rich content. To this end, we develop a two-stage training scheme and curate a dataset, including four parts: prompt refinement, similar image generation, inpainting \& outpainting, and instruction-based editing. For the first stage, we train MLLM to understand the properties of image generation and editing, enabling it to give fitting responses. For the second stage, we optimize the SD unet to adapt to the generation prompt. Moreover, we propose a DM-based reference restoration network (DiffRIR) to mitigate disparities in texture, contrast, and brightness for image editing. Consequently, LLMGA can offer design suggestions and enhance results based on user's requests during interactions.

\clearpage  

\title{Supplemental Material: \\LLMGA: Multimodal Large Language Model based Generation Assistant}

\author{	Bin Xia\inst{1}, Shiyin Wang\inst{2}, Yingfan Tao\inst{2}, Yitong Wang\inst{2}, and Jiaya Jia\inst{1} }
\institute{ The Chinese University of Hong Kong \and ByteDance Inc \\\href{https://llmga.github.io/}{https://llmga.github.io/}}

\maketitle

\section{Overview}
The overview of the supplementary materials:

\textbf{(1)} 
We have provided implementation details for the LLMGA (Sec.~\ref{sec:implement}).

\textbf{(2)} We offer a comprehensive introduction to the creation of the training dataset (Sec.~\ref{sec:sup-data}).

\textbf{(3)} We discussed the differences between BLIP-diffusion and LLMGA  (Sec.~\ref{sec:blip}).

\textbf{(4)} We provided the VQA performance of LLMGA (Sec.~\ref{sec:vqa}).

\textbf{(5)} We have provided a detailed introduction on how to use LLMGA to implement instruction-based image editing. (Sec.~\ref{sec:sup-editing_details})

\textbf{(6)} Additional details of the training and evaluation for LLMGA are elaborated in this part. Furthermore, a detailed training process for Stable Diffusion XL (SDXL) is also included (Sec.~\ref{sec:sup-train_details}).

\textbf{(7)} We present more comparison details on LLMGA and LLMGA Embedding, along with corresponding analyses (Sec.~\ref{sec:sup-control_scheme}).

\textbf{(8)} 
We show the LLMGA's integrated image and text picture book creation and design capabilities  (Sec.~\ref{sec:demo}).

\textbf{(9)} More visual results are showcased to highlight LLMGA's performance in T2I generation (Sec.~\ref{sec:sup-t2i}). 

\textbf{(10)} More visual results are provided in instruction-based image editing (Sec.~\ref{sec:sup-iedit}). 

\textbf{(11)} More visual results are presented for LLMGA on inpainting and outpainting (Sec.~\ref{sec:sup-inpainting}).

\textbf{(12)} More Visual results effectively demonstrate the prowess of DiffRIR in addressing brightness and contrast disparities between newly generated and retained regions during image editing, along with its ability to enhance texture details (Sec.~\ref{sec:sup-restoration}).

\textbf{(13)} We provide additional examples showcasing the interactive generation and editing capabilities of LLMGA (with SDXL) (Sec.~\ref{sec:sup-sdxl}).

\section{Implementation Details}
\label{sec:implement}
For the first stage of training, we employ the pretrained LLaVA-1.5-7B or LLaVA-1.5-13B as the initial MLLM. We utilize the AdamW optimizer, setting the learning rate to $2\times10^{-5}$. Moreover, we adopt CosineLR as the learning rate scheduler. The batch size per device and epochs are set to $16$ and $1$, respectively. Besides, the training ratios for VQA (LLaVA v1.5 mix665k), QA (including general QA, all kinds of design, picture book generation, and illustration generation), prompt refinement, similar image generation, inpainting \& outpainting, and instruction-based editing are specified as $1$, $0.3$, $0.3$, $0.3$, $0.3$ and $0.3$, respectively.

For the second stage of training, we adopt the Stable Diffusion 1.5 (SD1.5) as the initial image generation or inpainting \& outpainting model. We train these models with the AdamW optimizer, setting the learning rate to $1\times10^{-5}$. The batch size is set to $32$. We train SD1.5 by $1\times10^{5}$ iterations.

For the restoration network, we train DiffRIR on DIV2K~\cite{DIV2K} and Flickr2K~\cite{Flickr2K} datasets with the same GAN-based loss function as DiffIR. The batch sizes are set to $64$, and the
LQ patch sizes are set to $64\times64$. We use Adam optimizer, setting the learning rate to $2\times10^{-4}$. We train this model by $4\times10^{5}$ iterations. 

\section{Data}
\label{sec:sup-data}
\begin{figure}[t]
	\centering
 \resizebox{0.7\linewidth}{!}{
	\includegraphics[height=4cm]{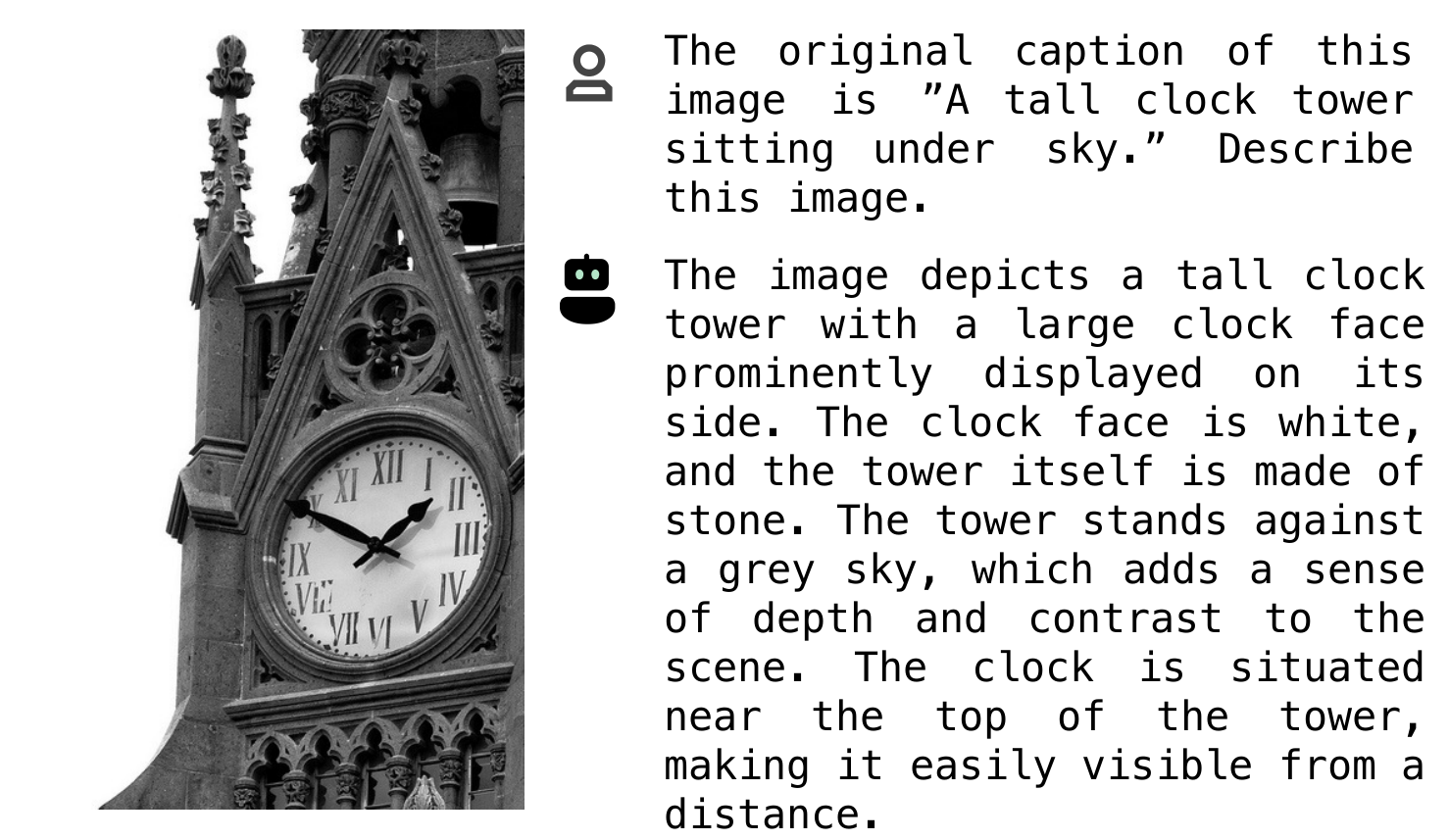}
 }
  \vspace{-2mm}
	\caption{An example of generating detailed descriptions for description data.}
	\label{fig:data}
 \vspace{-2mm}
\end{figure}

For the first stage of training, we constructed a training dataset that requires detailed descriptions of images to assist the LLM in better understanding the compositional details of images and supporting image generation and editing. Specifically, we utilized the MSCOCO~\cite{mscoco} datasets, encompassing rich real-world scenarios. Then, we employed GPT4-V to generate detailed and visually compelling descriptions for these datasets. The prompt format for GPT4-V to generate a detailed description is as follows: \textit{"The original caption of this image is 'ORIGINAL CAPTION'. Describe this image."}, where \textit{'ORIGINAL CAPTION'} is a short caption provided by the dataset. 
An example is shown in Fig.~\ref{fig:data}.

\begin{figure}[h]
	\centering
 \resizebox{0.9\linewidth}{!}{
	\includegraphics[height=4cm]{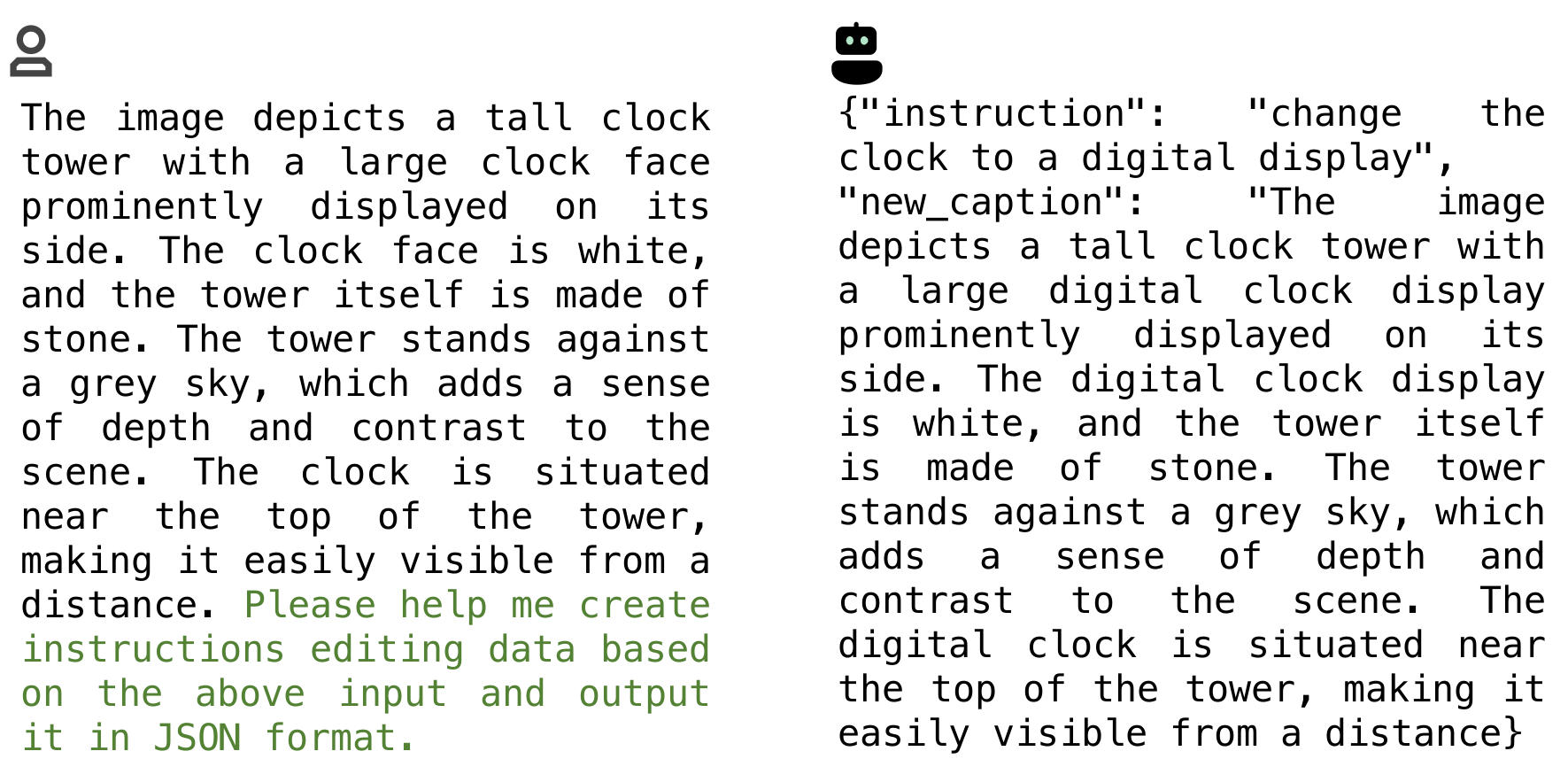}
 }
  \vspace{-2mm}
	\caption{An example of generating detailed descriptions for instruction-based editing data.}
	\label{fig:data2}
 \vspace{-2mm}
\end{figure}

In order to better adapt the LLMGA to instruction-based image editing, we finetuned Mixtral7Bx8 using short instruction editing training data from InstructPix2Pix. This enabled it to learn how to generate corresponding editing data and target descriptions in JSON format based on detailed descriptions. Then, we used the finetuned Mixtral7Bx8 to create corresponding editing data for detailed descriptions of MSCOCO images made with GPT4-V. An example is shown in Fig.~\ref{fig:data2}. After that, we filtered out data with incorrect output formats to obtain the final instruction-based editing training dataset.

Furthermore, we have gathered a vast amount of design books and graphic story datasets for creating parts QA data, which has been used to train the LLMGA. This enables the LLMGA to excel in various design tasks, including logo design, fashion design, infographic creation, and poster design. Additionally, it can generate illustrated storybooks that intertwine text and images, tailored to user specifications.

\section{Discuss BLIP-diffusion and LLMGA.}
\label{sec:blip}
\begin{figure}[htb]

	\centering
	\resizebox{1\linewidth}{!}{
	\includegraphics{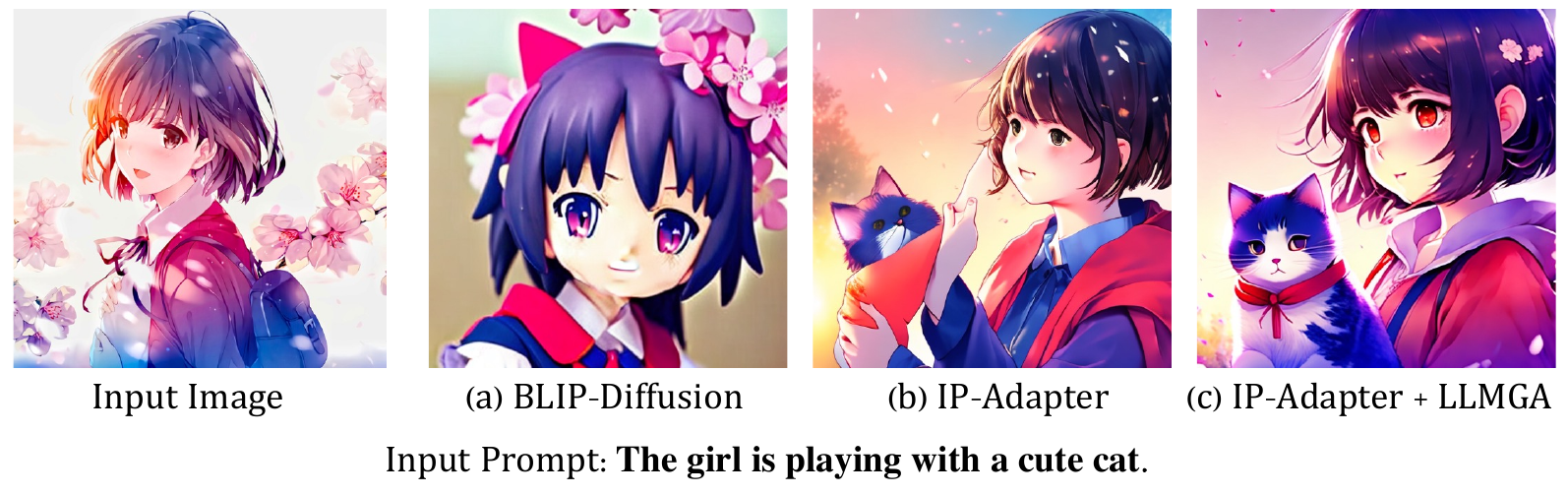} 
	}

	\caption{The comparison on subject-driven generation.}

	\label{fig:blip}
\end{figure}

\textbf{(1)} Unlike works as BLIP-diffusion~\cite{blip-diffusion}, which uses MLLMs primarily as image and text encoders, our LLMGA aims to retain the multi-turn conversation and general cognitive ability of LLMs, acting as a designer and assistant to produce satisfactory images through conversation. \textbf{(2)} Subject-driven generation can be implemented with IP-Adapters, easily integrated into our SD-ft. Fig.~\ref{fig:blip} shows our LLMGA achieves better outcomes. \textbf{(3)} We will add the discussion of these methods in our paper.

\section{VQA Evaluation.}
\label{sec:vqa}
 In Tab.~\ref{tab:vqa}, compared to other MLLMs with image generation ability, our LLMGA has a much better performance on both VQA and image generation. 

\begin{table}[t]
\centering
 \caption{VQA comparison.}
  \resizebox{0.9\linewidth}{!}{
    \begin{tabular}{c|c|ccc}
    \toprule
    Method & Can image generation? & MMB$\uparrow$ & VQA$^{V2}$$\uparrow$ & POPE$\uparrow$ \\
    \midrule
    LLaVA1.5-7b & \XSolidBrush     & 64.3  & 78.5  & 85.9 \\
    \midrule
    GILL  & \Checkmark     & 4.8  & 20.4  & 50.1 \\
    LLMGA-embedding & \Checkmark     & 35.3  & 56.3  & 67.4 \\
    LLMGA-7b (Ours) & \Checkmark     & \textbf{62.9}  & \textbf{77.2}  & \textbf{83.8} \\
    \bottomrule
    \end{tabular}%
    }

  \label{tab:vqa}%
\end{table}%

\section{More Details on Instruction-based Editing}
\label{sec:sup-editing_details}

Given an image and corresponding editing instructions, our LLMGA provides a description of the image and a target description after editing based on the given editing instructions. We then perform a Direct Inversion~\cite{directinversion} on the image to obtain an initial noise map. Utilizing this noise map, along with the original image's description and the edited target description, we employ the straightforward prompt-to-prompt~\cite{prompt2prompt} method to achieve the desired edited outcome.

\section{More Training and Evaluation Details}
\label{sec:sup-train_details}
In addition to the LLMGA (SD1.5) described in the paper, we have also conducted training for LLMGA (SDXL). LLMGA (SDXL) follows a training process and configuration similar to that of LLMGA (SD1.5) but demonstrates even higher-quality generation. Specifically, for the first training stage, we employ the same MLLM as utilized in LLMGA (SD1.5). Regarding MLLM, we utilize the pretrained LLaVA-1.5-7B or LLaVA-1.5-13B as the initial MLLM. Our optimization approach involves the use of the AdamW optimizer, with the learning rate set at $2\times10^{-5}$ and weight decay at $0$, respectively. 
Additionally, we adopt the CosineLR learning rate scheduler. The total batch size and number of epochs are configured at $128$ and $1$, respectively. MLLM training is carried out on our constructed datasets as described in the paper. Furthermore, the training ratios for VQA (LLaVA v1.5 mix665k), QA (Alpaca), prompt refinement, similar image generation, inpainting \& outpainting, and instruction-based editing are specified as $1$, $0.3$, $0.3$, $0.3$, $0.3$ and $0.3$, respectively.

For the second stage of training, we adopt the Stable Diffusion XL (SDXL)~\cite{sdxl} as the initial image generation or editing model. We train these models with AdamW optimizer, setting the learning rate to $1\times10^{-5}$. The total batch size is set to $32$. We train SDXL by $5\times10^{4}$ iterations. 
For both T2I generation and inpainting \& outpainting, we conduct training on the LAION-Aesthetic dataset, using the generation prompts generated by the first-stage pretrained MLLM as guidance.  The input patch sizes are set to $1024 \times 1024$. Moreover, similar to SD1.5, we randomly generate masks for SDXL inpainting \& outpainting, including box masks, irregular masks, and boundary masks.

\section{More Details on Control Scheme}
\label{sec:sup-control_scheme}
\begin{figure}[t]
	\centering
 \resizebox{0.7\linewidth}{!}{
	\includegraphics[height=4cm]{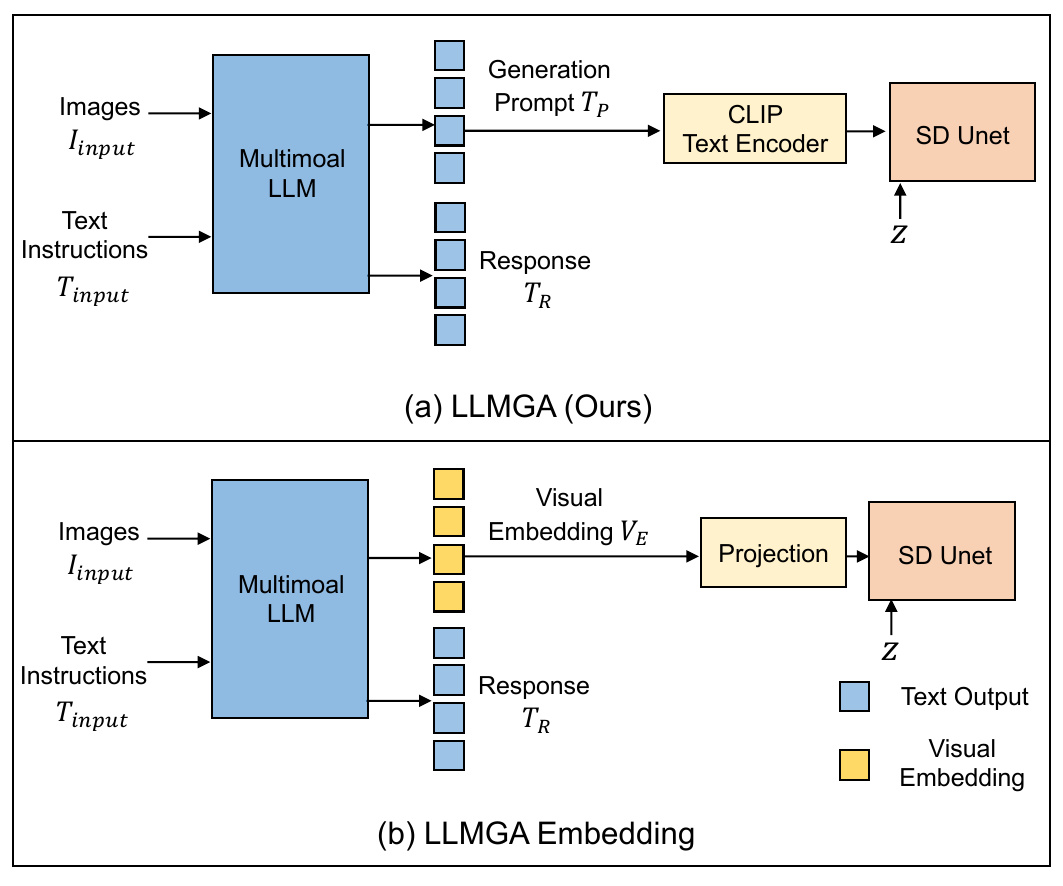}
 }
  \vspace{-3mm}
	\caption{The illustration of LLMGA and LLMGA Embedding. LLMGA uses language generation prompts as a control scheme, while LLMGA Embedding uses visual embedding as a control scheme. }
	\label{fig:control_sup}
 \vspace{-5mm}
\end{figure}

The form in which to establish a control link between MLLM and SD is a question that requires careful consideration. In this paper, as illustrated in Fig.~\ref{fig:control_sup}, we explore two control schemes: namely, our adopted language-based generation prompt control (Fig.\ref{fig:control_sup} (a)) and visual embedding-based control scheme (Fig.\ref{fig:control_sup} (b)). Here, we will provide a detailed overview of the training approach for LLMGA Embedding, followed by a comparison and analysis of these schemes.

For the training of LLMGA Embedding, we employ a comprehensive three-stage training scheme. \textbf{(1)} In the first stage, for $T_R$, we use the same auto-regressive cross-entropy loss ($\mathcal{L}_{MLLM}$, Eq.~\ref{eq:loss_mllm}) as LLMGA, and for visual embedding, we utilize the visual embedding loss ($\mathcal{L}_{embed}$, Eq.~\ref{eq:loss_embed}). We simultaneously apply the same training settings as LLMGA.  \textbf{(2)} In the second stage, we initiate joint optimization of MLLM and SD. Unlike LLMGA, in this phase, we optimize only the projection layer for LLMGA embedding, using the SD loss ($\mathcal{L}_{SD}$, Eq.~\ref{eq:loss_SD}), and then freeze the parameters of Unet and MLLM. \textbf{(3)} In the third training stage, we freeze the parameters of MLLM and projection, and then optimize the SD Unet using the SD loss ($\mathcal{L}_{SD}$, Eq.~\ref{eq:loss_SD}).
\begin{equation}
\label{eq:loss_mllm}
\mathcal{L}_{MLLM}=\mathbf{C E}\left(\mathbf{T}_{R}, \mathbf{T}_{GT}\right),
\end{equation}
\begin{equation}
\label{eq:loss_embed}
\mathcal{L}_{embed}=\left\|\phi_{proj}(\mathbf{V}_E)-\phi_{CLIP}(\mathbf{T}_{caption})\right\|_2^2,
\end{equation}
\begin{equation}
\label{eq:loss_SD}
\vspace{-1mm}
\mathcal{L}_{SD}=\mathbb{E}_{\mathbf{Z}_t, \mathbf{C}, \epsilon, t}\left(\left\|\epsilon-\epsilon_\theta\left(\mathbf{Z}_t, \mathbf{C}\right)\right\|_2^2\right),
\end{equation}
where $\mathbf{T}_{R}$ denotes the generated text response, and $\mathbf{T}_{GT}$ represents the ground-truth target. $\mathbf{CE}(.)$ signifies auto-regressive cross-entropy. $\phi_{proj}(.)$ denotes the projection involving three linear layers. $\phi_{CLIP}(.)$ is the CLIP text encoder. $\mathbf{V}_E$ stands for visual embedding, and $\mathbf{T}{caption}$ corresponds to the original caption from the datasets. $\mathcal{L}_{SD}$ represents the diffusion loss as described in the paper.

The results are presented in Fig.~\textbf{\textcolor{blue}{6}} in the main paper. Despite comprehensive training, we find that the embedding-based approaches still lags behind LLMGA. Moreover, with an increase in the number of dialogue turns, there is a noticeable decrease for embedding-based approaches in both the accuracy and quality of generation. 
This observation can be well comprehended through the processing mechanism of LLM. Specifically, LLM predicts an embedding based on previous input images and texts (Eq.~\ref{eq:llm_embed}). After that, embedding is refined through a linear layer, categorizing it into a fixed language domain (Eq.~\ref{eq:llm_lang}). 
This is because predicted embeddings exist in a continuous space, inherently imprecise and filled with noise. For instance, the same embedding, depending on the sampling probabilities, may generate different semantic words, indicating that embeddings are filled with various forms of noise. Mapping embeddings to a fixed and specific language domain by classification effectively eliminates this noise, enabling precise control over SD generation. The decline in performance of embedding-based methods with an increase in conversation turns is attributed to the introduction of additional noise into the predicted embedding as more prior conversation information is incorporated, thereby affecting accuracy.

\begin{equation}
\label{eq:llm_embed}
\mathbf{E}=\Psi_{body}(\mathbf{I}_{input},\mathbf{T}_{input}),
\end{equation}
\begin{equation}
\label{eq:llm_lang}
\mathbf{T}=\Psi_{linear}(\mathbf{E}),
\end{equation}
where $\mathbf{I}_{input}$ and $\mathbf{T}_{input}$ represent all preceding input images and texts. $\Psi_{body}$ signifies the network body of LLM, producing an embedding $\mathbf{E}$. $\Psi_{linear}$ constitutes the final linear layer in LLM, tasked with categorizing the embedding into a predetermined text domain, resulting in the generation of text $\mathbf{T}$.

In summary, compared with embedding based methods, our LLMGA using detailed language prompts for control generation has the following advantages:
 \begin{itemize}
 \item  The embeddings predicted by the MLLM are often filled with noise. This can be filtered out by mapping them to a fixed language domain, enabling precise control of SD.
 \item  Detailed language prompts can make the network more transparent and interactive, allowing users to understand MLLM's thoughts for generating images.  
 \item   MLLM is pre-trained on vast textual datasets. Explicit language prompts rather than implicit embeddings are more advantageous for MLLM to generate prompts and comprehend context.
 \item  Dynamic-sized language prompt facilitates the addition of generation requests during interactions.
 \item  Training is more simple and more efficient.
\end{itemize}

\section{Showcasing LLMGA's Integrated Image and Text Picture Book Creation and Design Capabilities}
\label{sec:demo}
 
\begin{figure*}[t]
	\centering
 \resizebox{1\linewidth}{!}{
	\includegraphics[height=4cm]{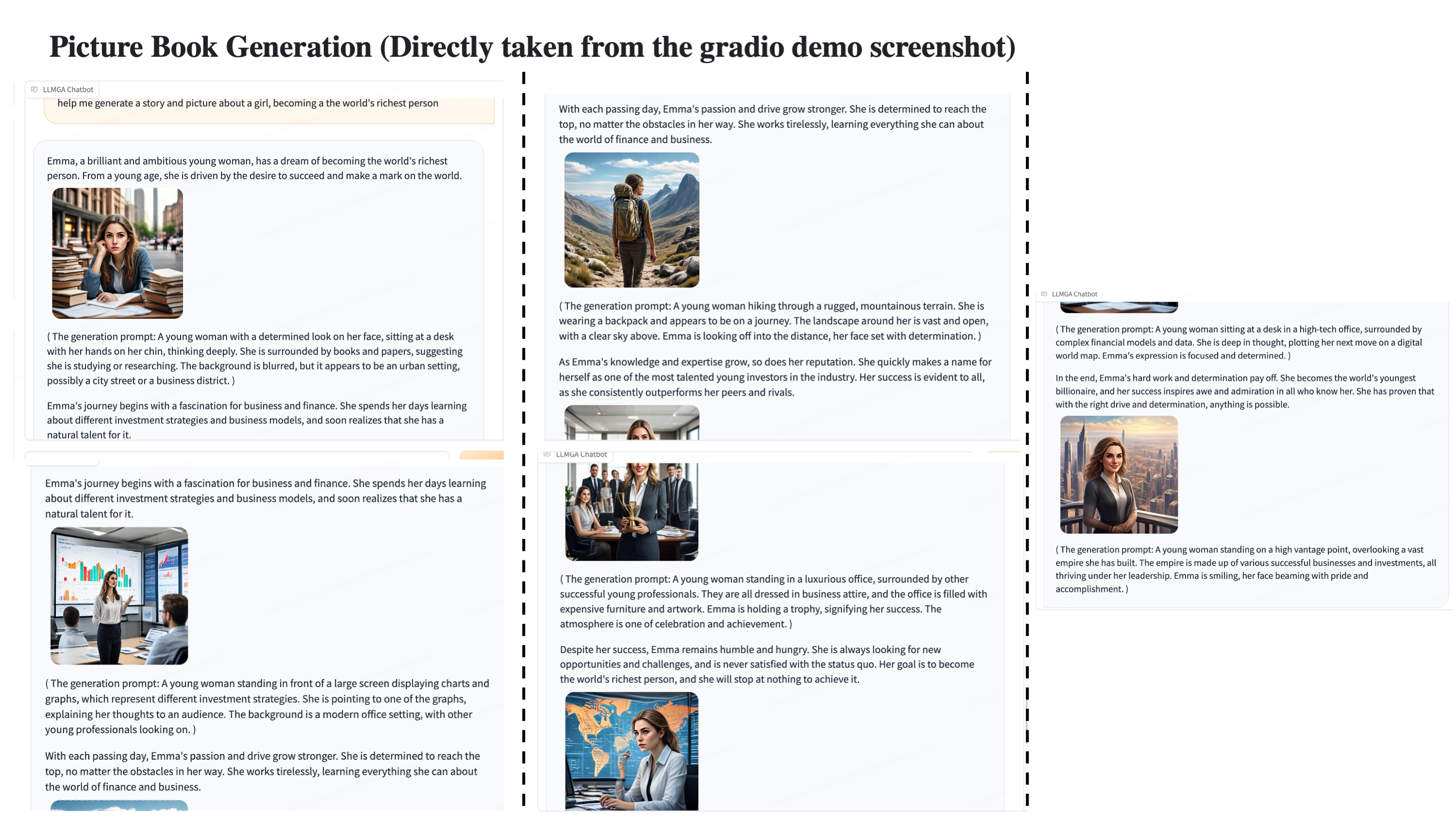}
 }
  \vspace{-3mm}
\caption{Showcase of LLMGA creating integrated image and text picture book. The image content is captured directly from our Gradio demo.}
	\label{fig:demo1}
\end{figure*}

As shown in Fig.~\ref{fig:demo1},  LLMGA can create integrated image and text picture book. With a single user's instruction, LLMGA can generate an interwoven storybook of text and illustrations. Moreover, LLMGA can perform illustration Generation. LLMGA can interactively generate story illustrations based on user-input story snippets.

\begin{figure*}[t]
	\centering
 \resizebox{1\linewidth}{!}{
	\includegraphics[height=4cm]{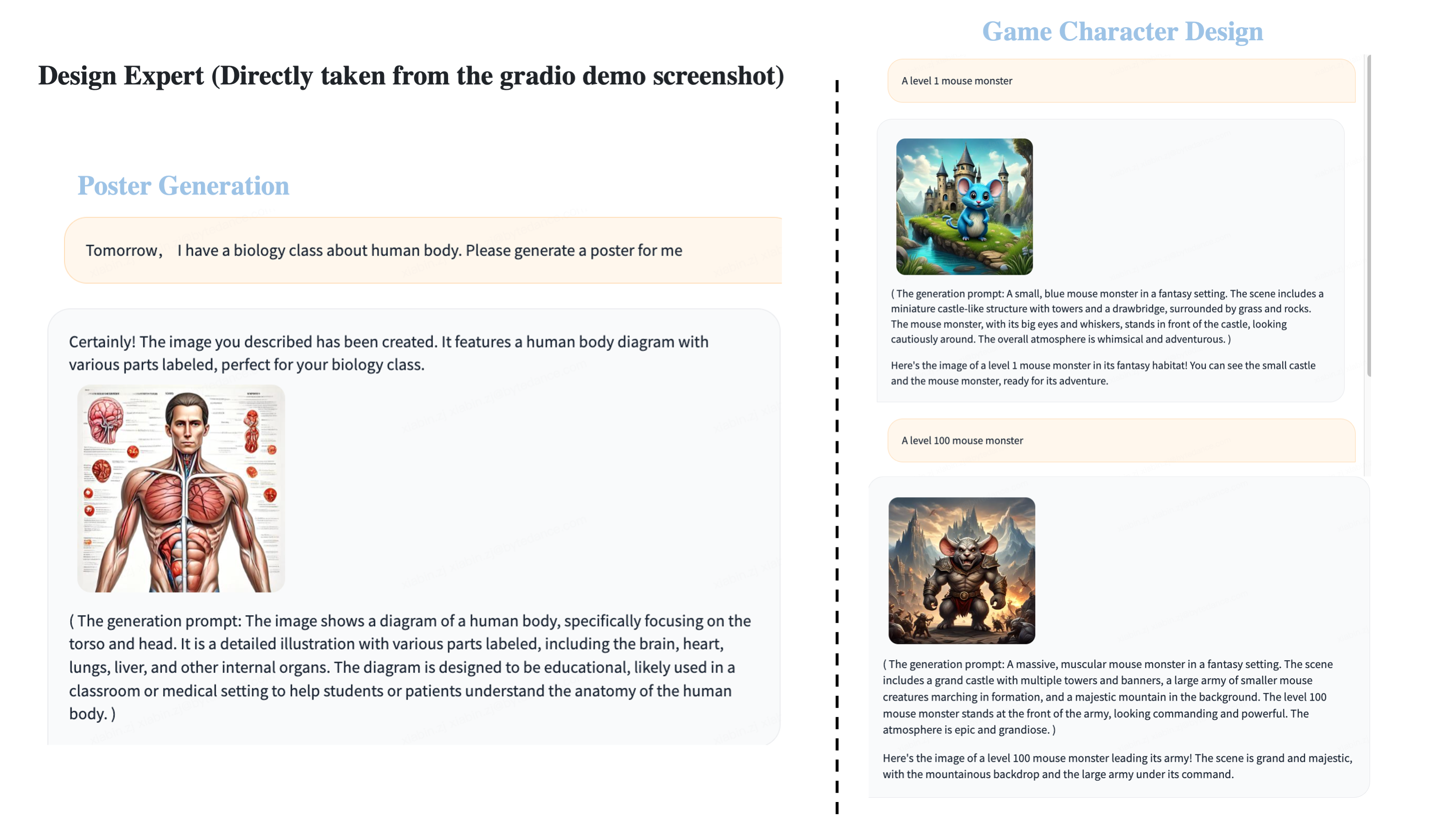}
 }
  \vspace{-3mm}
\caption{Showcase of LLMGA assisting image designing. The image content is captured directly from our Gradio demo.}
	\label{fig:demo2}
\end{figure*}

As shown in Fig.~\ref{fig:demo2}, LLMGA is a design expert. LLMGA incorporates an extensive array of image design data, offering deep insights for a wide range of design tasks, including logo creation, game character design, poster design, T-shirt design, infographic design, and more.

\section{More Results on T2I Generation}
\label{sec:sup-t2i}
\begin{figure*}[t]
	\centering
 \resizebox{1\linewidth}{!}{
	\includegraphics[height=4cm]{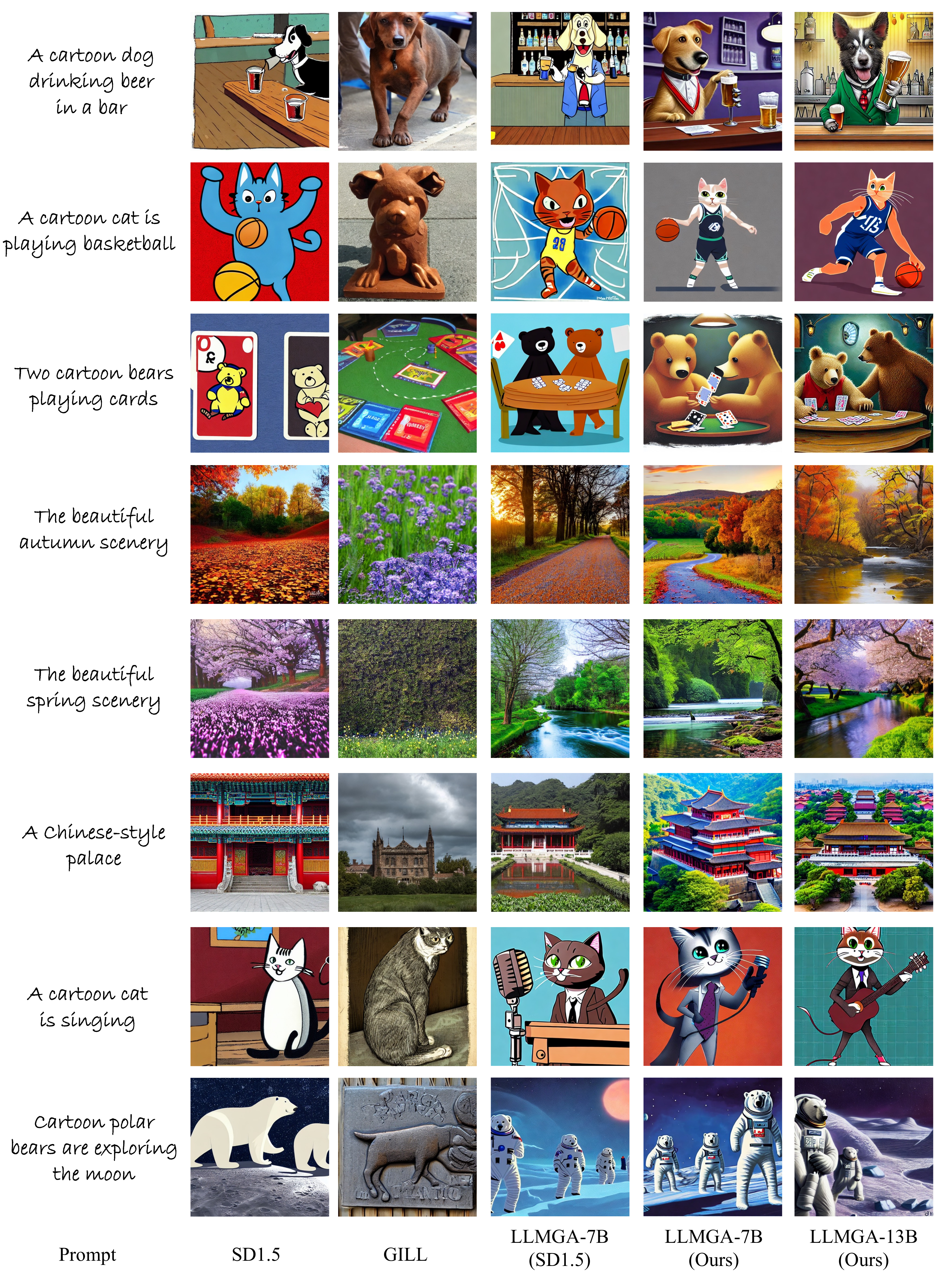}
 }
  \vspace{-8mm}
\caption{Visual comparison on \textbf{T2I}. LLMGA can refine short prompts by adding details, such as clothing, background, and actions. In addition, for unfamiliar concepts, LLMGA can utilize its extensive knowledge base from LLM to realize accurate generation.}
	\label{fig:t2i-sup}
\end{figure*}

The more T2I generation visual results are shown in Fig.~\ref{fig:t2i-sup}. LLMGA can leverage its vast comprehension, reasoning abilities, and knowledge reservoir to generate visuals with more details.  Furthermore, LLMGA refine prompts based on user requirements.

\section{More Results on Instruction-based Editing}
\label{sec:sup-iedit}
\begin{figure*}[t]
	\centering
 \resizebox{1\linewidth}{!}{
	\includegraphics[height=4cm]{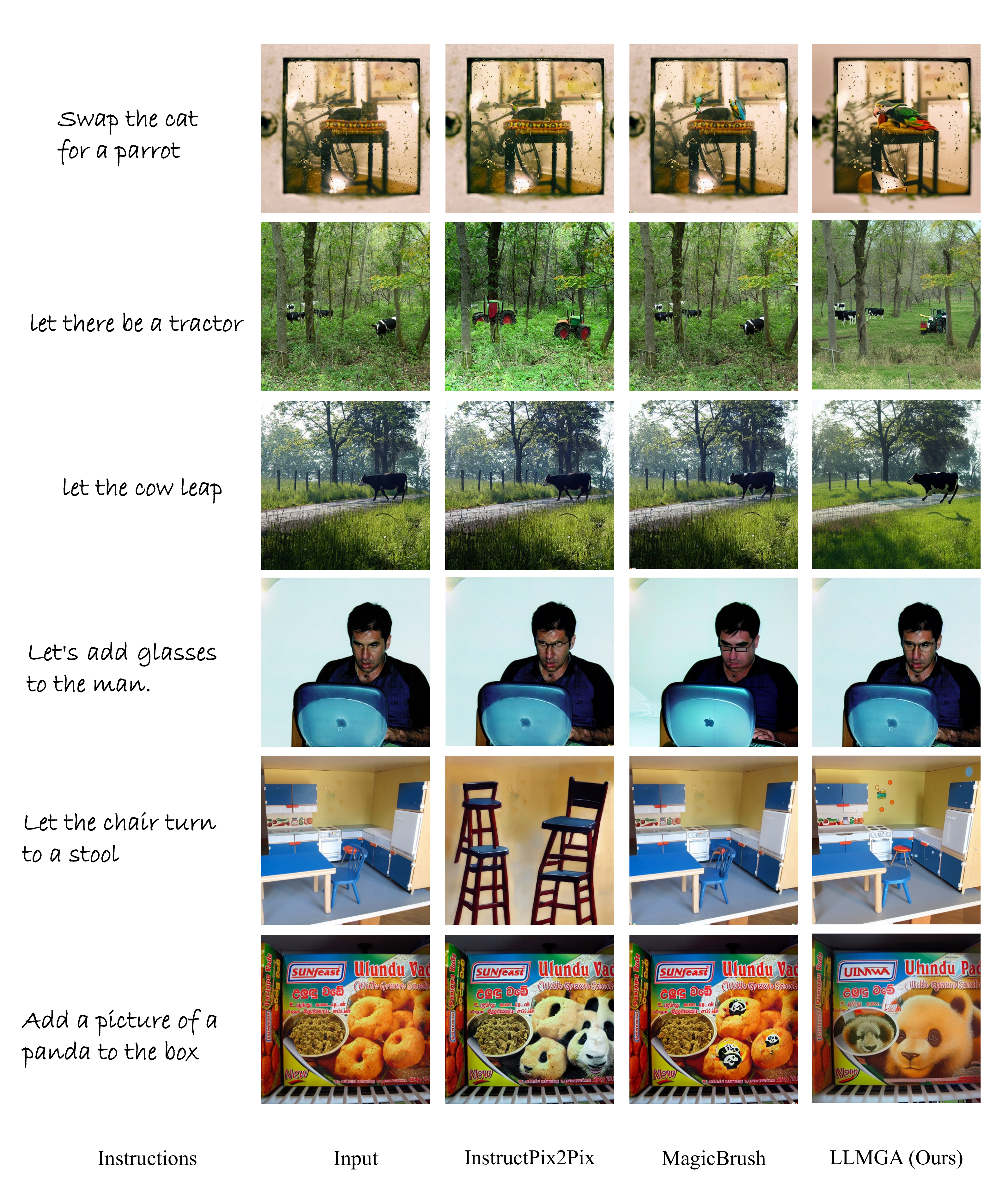}
 }
  \vspace{-8mm}
	\caption{Visual comparison on \textbf{instruction-based editing}.   }
	\label{fig:edit-sup}
\end{figure*}

The more instruction-based editing visual results are depicted in Fig.~\ref{fig:edit-sup}. Our LLMGA can realize more accurate and visually pleasing editing according to the requirements of users.

\section{More Results on Inpainting and Outpainting}
\label{sec:sup-inpainting}
\begin{figure*}[t]
	\centering
 \resizebox{1\linewidth}{!}{
	\includegraphics[height=4cm]{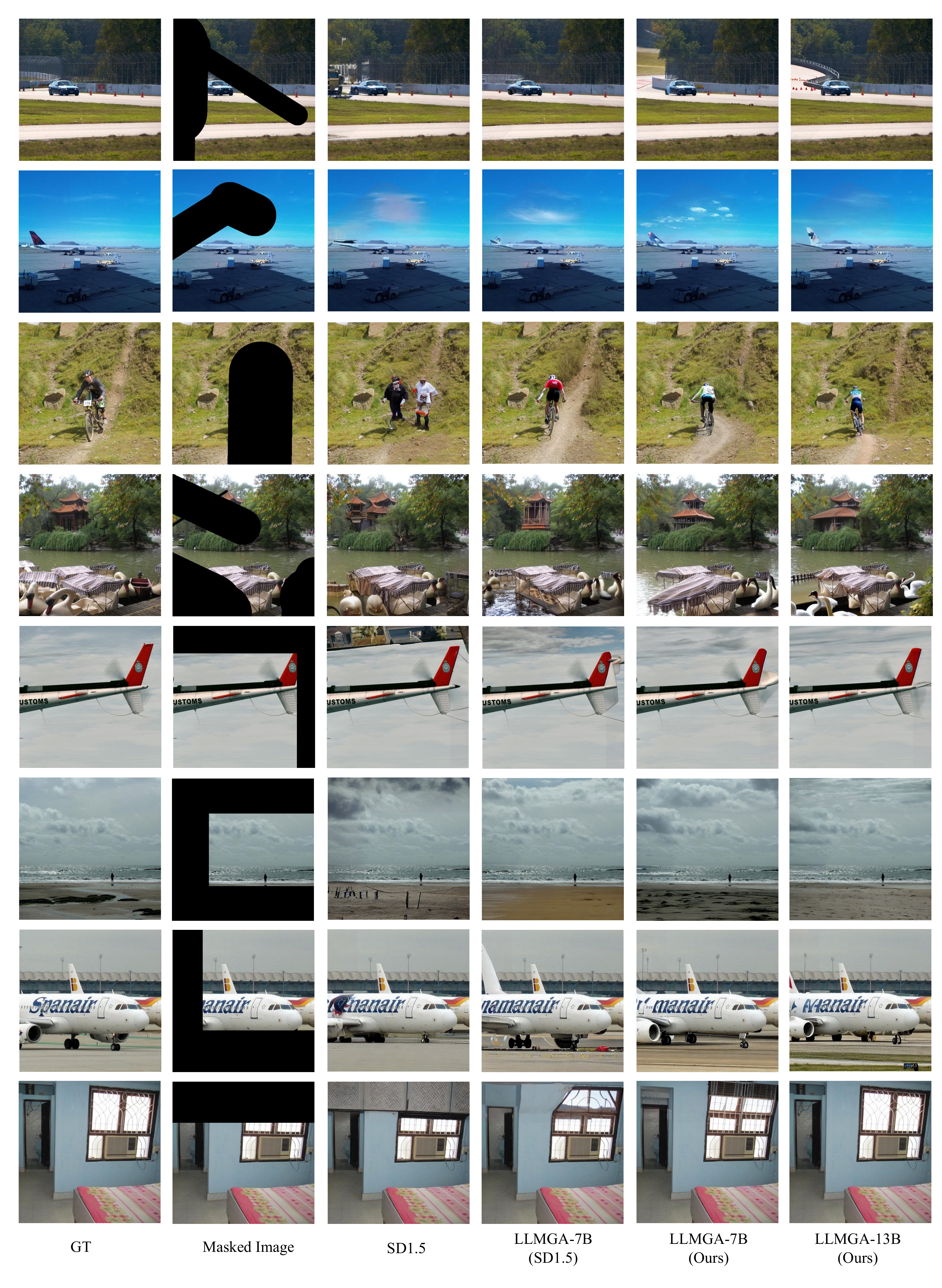}
 }
  \vspace{-8mm}
	\caption{Visual comparison on \textbf{inpainting and outpainting}. LLMGA can infer complete images based on input masked images.  }
	\label{fig:painting-sup}
 \vspace{-4mm}
\end{figure*}

The more inpainting and outpainting visual results are depicted in Fig.~\ref{fig:painting-sup}. LLMGA can harness its extensive comprehension, reasoning abilities, and knowledge reservoir to infer plausible complete images based on given masked images.  Additionally, as demonstrated in paper Fig.~\textbf{\textcolor{blue}{1}}, LLMGA can edit images in accordance with user specifications and masked images.

\section{More Results on Image Restoration}
\label{sec:sup-restoration}
\begin{figure*}[t]
	\centering
 \resizebox{1\linewidth}{!}{
	\includegraphics[height=4cm]{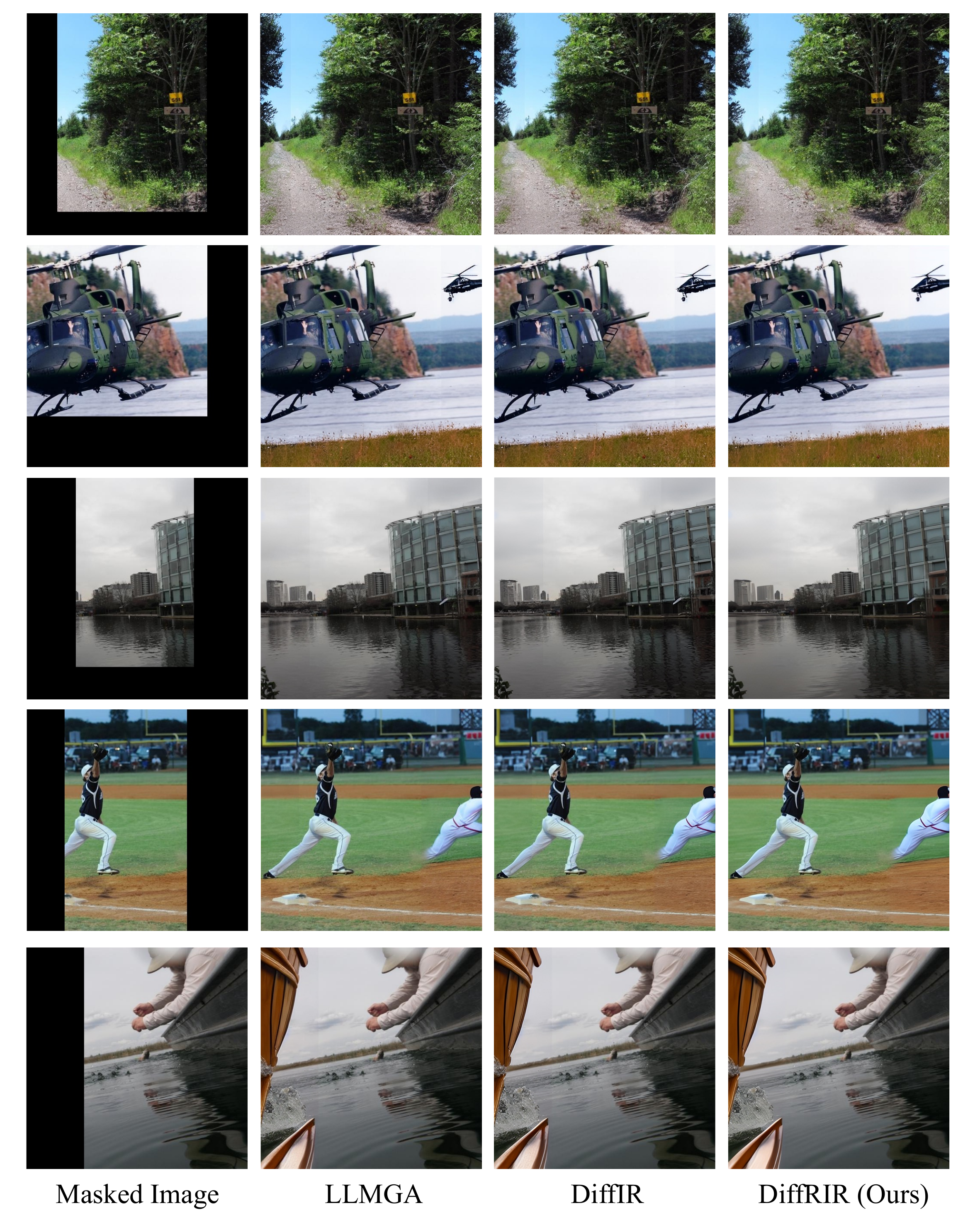}
 }
  \vspace{-8mm}
	\caption{More visual comparison on image \textbf{restoration}. }
	\label{fig:IR-sup}
 \vspace{-4mm}
\end{figure*}

The more image restoration results are shown in Fig.~\ref{fig:IR-sup}. Our DiffRIR can alleviate the texture, brightness, and contrast discrepancies, and generate more realistic details.

\section{More Results on Interactive Generation and Editing}
\label{sec:sup-sdxl}

\begin{figure*}[t]
	\centering
 \resizebox{1\linewidth}{!}{
	\includegraphics[height=4cm]{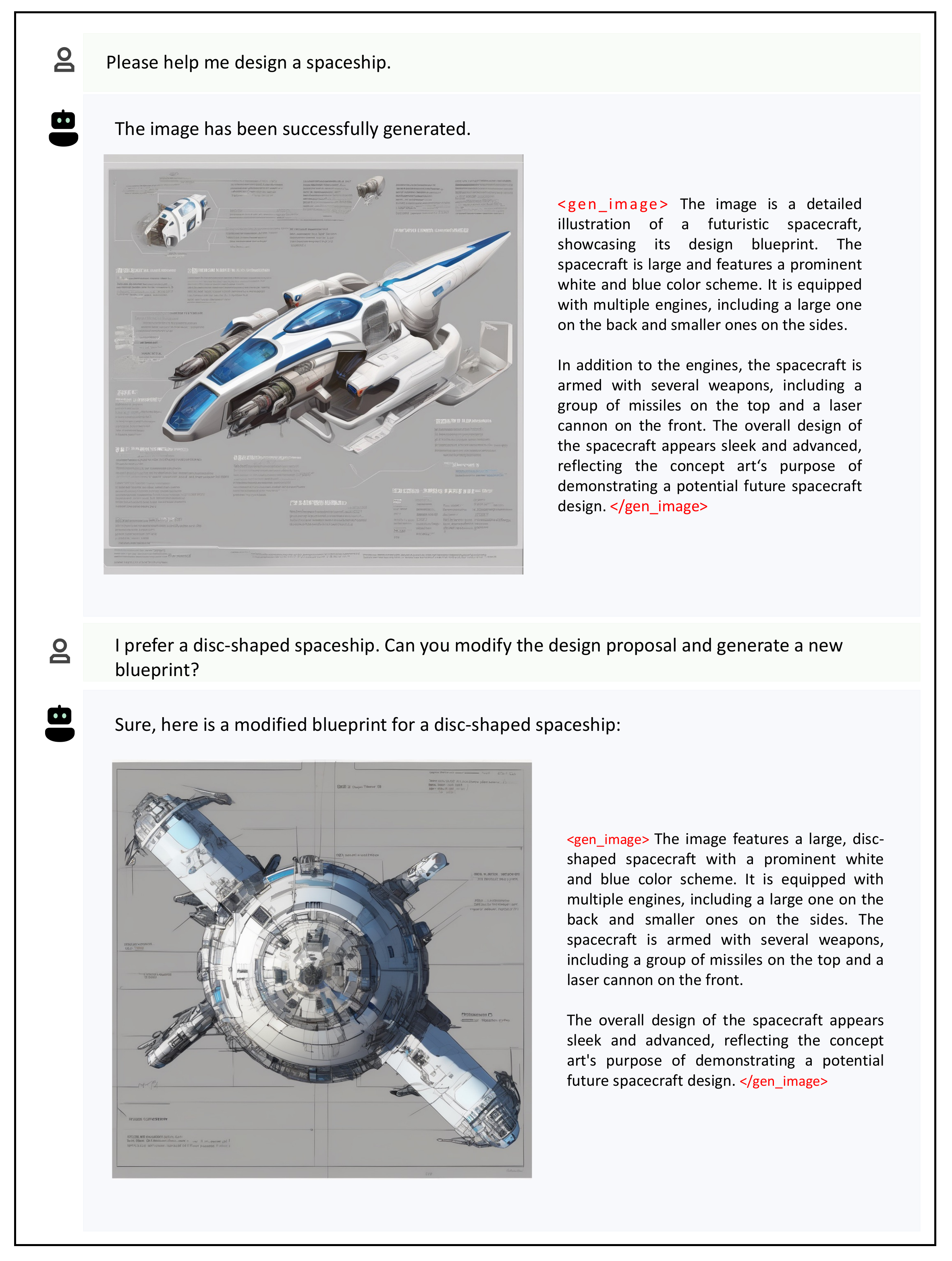}
 }
  \vspace{-8mm}
	\caption{Examples of LLMGA on interactive generation and editing. }
	\label{fig:poster1-sup}
 \vspace{-4mm}
\end{figure*}

\begin{figure*}[t]
	\centering
 \resizebox{1\linewidth}{!}{
	\includegraphics[height=4cm]{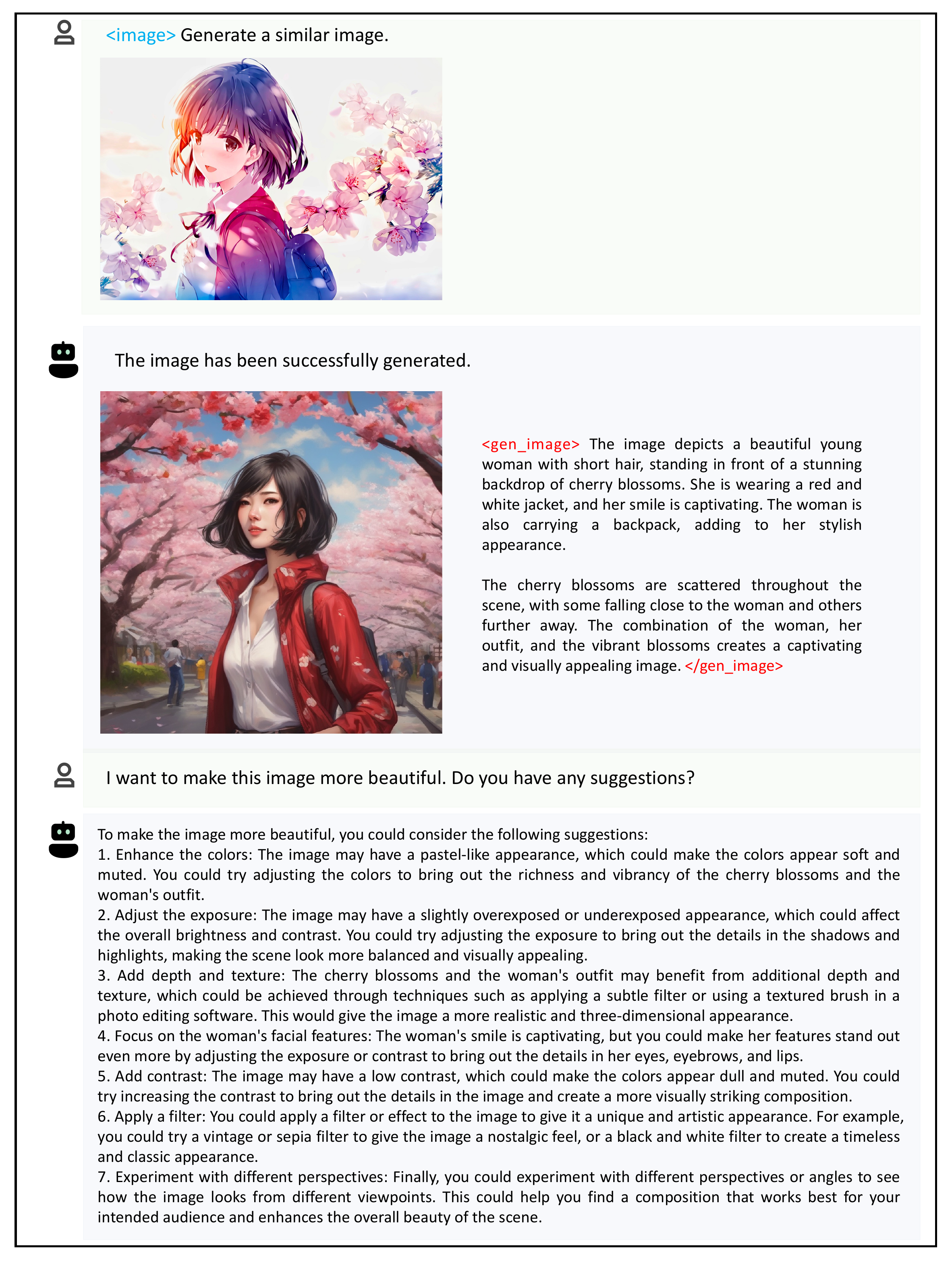}
 }
  \vspace{-8mm}
	\caption{Examples of LLMGA on interactive generation and editing. }
	\label{fig:poster2-sup}
 \vspace{-4mm}
\end{figure*}

\begin{figure*}[t]
	\centering
 \resizebox{1\linewidth}{!}{
	\includegraphics[height=4cm]{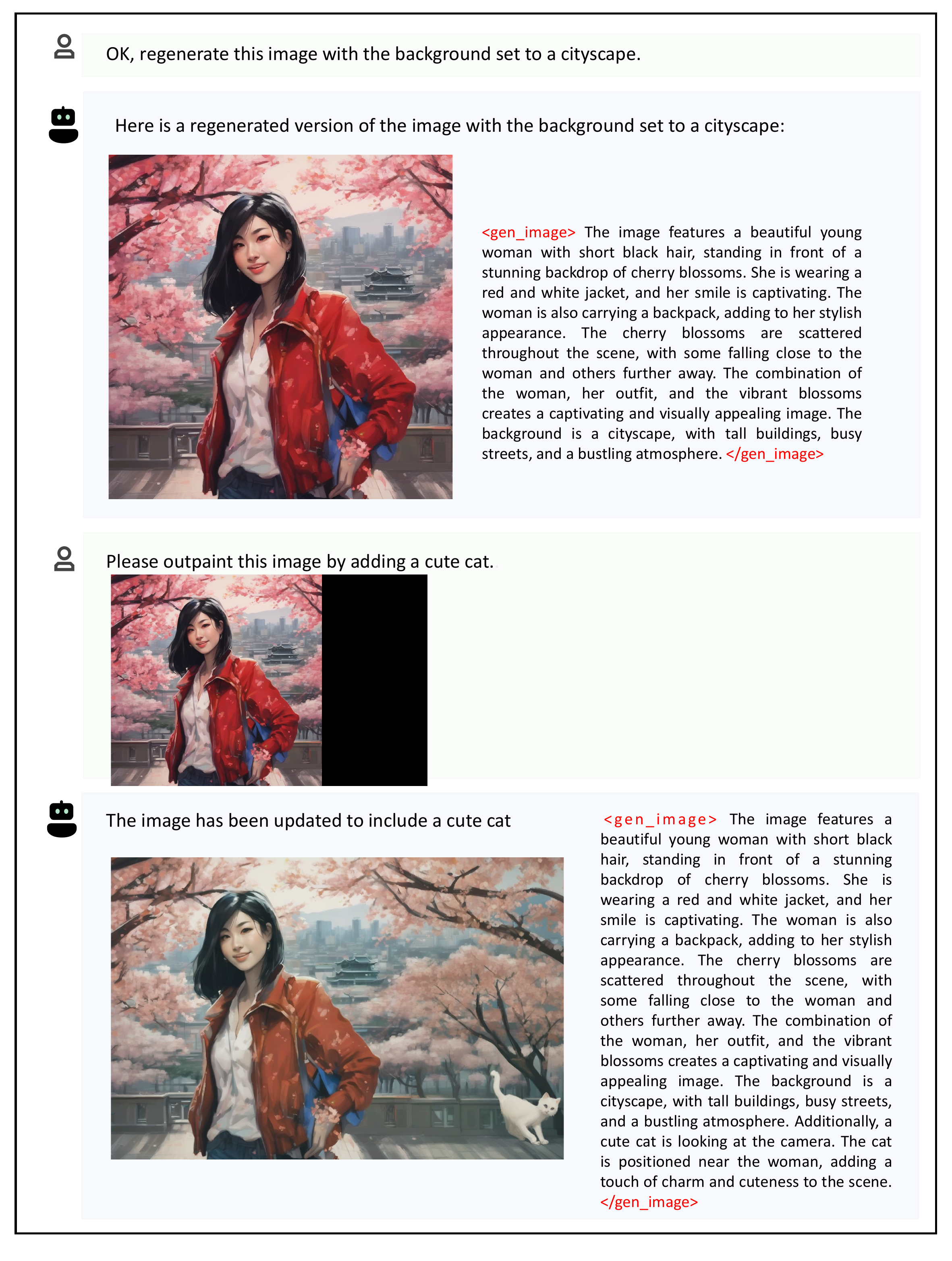}
 }
  \vspace{-8mm}
	\caption{Examples of LLMGA on interactive generation and editing. }
	\label{fig:poster3-sup}
 \vspace{-4mm}
\end{figure*}

The results are shown in Figs.~\ref{fig:poster1-sup},~\ref{fig:poster2-sup}, and~\ref{fig:poster3-sup}. The results presented here were obtained using LLMGA-7b (SDXL-ft). It can be observed that LLMGA, leveraging the understanding, reasoning abilities, and extensive knowledge repository of LLM, effectively assists users in image design in an interactive manner.

\clearpage
%
%
\bibliographystyle{splncs04}
\bibliography{main}
\end{document}